\documentclass{article}
\usepackage[dblblindworkshop, final]{neurips_2026}
\workshoptitle{AgenticOS: Co-designing Systems and ML Foundations of an OS Layer for Agentic AI}

\usepackage[utf8]{inputenc}
\usepackage[T1]{fontenc}
\usepackage{amsmath,amssymb}
\usepackage{booktabs}
\usepackage{microtype}
\usepackage{graphicx}
\usepackage{xcolor}
\usepackage{url}
\usepackage{hyperref}
\usepackage{multirow}
\usepackage{booktabs}
\usepackage{pifont}
\usepackage{listings}
\usepackage{todonotes}
\usepackage{svg}
\usepackage{subcaption}
\usepackage{adjustbox}
\usepackage[table]{xcolor} 

\usepackage[most]{tcolorbox}
\lstdefinestyle{promptstyle}{
    basicstyle=\ttfamily\scriptsize,
    breaklines=true,
    breakatwhitespace=true,
    columns=fullflexible,
    keepspaces=true,
    showstringspaces=false,
    aboveskip=0pt,
    belowskip=0pt
}
\newcommand{\cmark}{\ding{51}}
\newcommand{\xmark}{\ding{55}}
\title{AndroidFlux: Evaluating Failure Recovery and Recovery Action Reward Modeling in Mobile-Use Multimodal Agents}

\vspace{-12pt}

\author{%
Gyubeum Lim\\Georgia Institute of Technology\\\texttt{glim36@gatech.edu}
\And
Yemo Koo\\Konkuk University\\\texttt{yemoyemo010831@konkuk.ac.kr}
\AND
Guhnwoo Jang\\Georgia Institute of Technology\\\texttt{gjang34@gatech.edu}
\And
Zsolt Kira\\Georgia Institute of Technology\\\texttt{zkira@gatech.edu}}

\begin{document}
\maketitle

\vspace{-12pt}
\begin{abstract}
\vspace{-8pt}

Mobile-use multimodal agents are typically evaluated by task success from a clean initial state, which does not isolate their ability to recover from intermediate errors. Because mobile environments change dynamically, a general mobile-use agent must complete tasks robustly even from unexpected states and in unseen environments. This requires knowing when and where an agent makes errors, and whether it can recover from errors and subsequent states. To evaluate this capability, we introduce \textbf{AndroidFlux}, a failure-recovery benchmark built on AndroidWorld. AndroidFlux restores recorded trajectories to controlled checkpoints before, at, and after the first recoverable error, so that different agents continue from the same device state and history. On these checkpoints, a complementary reward-modeling benchmark evaluates recovery-action preferences using downstream rollout outcomes. We further release an offline reinforcement learning (RL) dataset that improves both error recovery and overall task success. Our evaluations show that recovery performance only partially correlates with clean-start success, and that explicit reasoning helps only models trained with GUI-domain specific RL.
\end{abstract}
\vspace{-8pt}

\begin{figure}[h]
\centering
\includegraphics[width=0.85\textwidth]{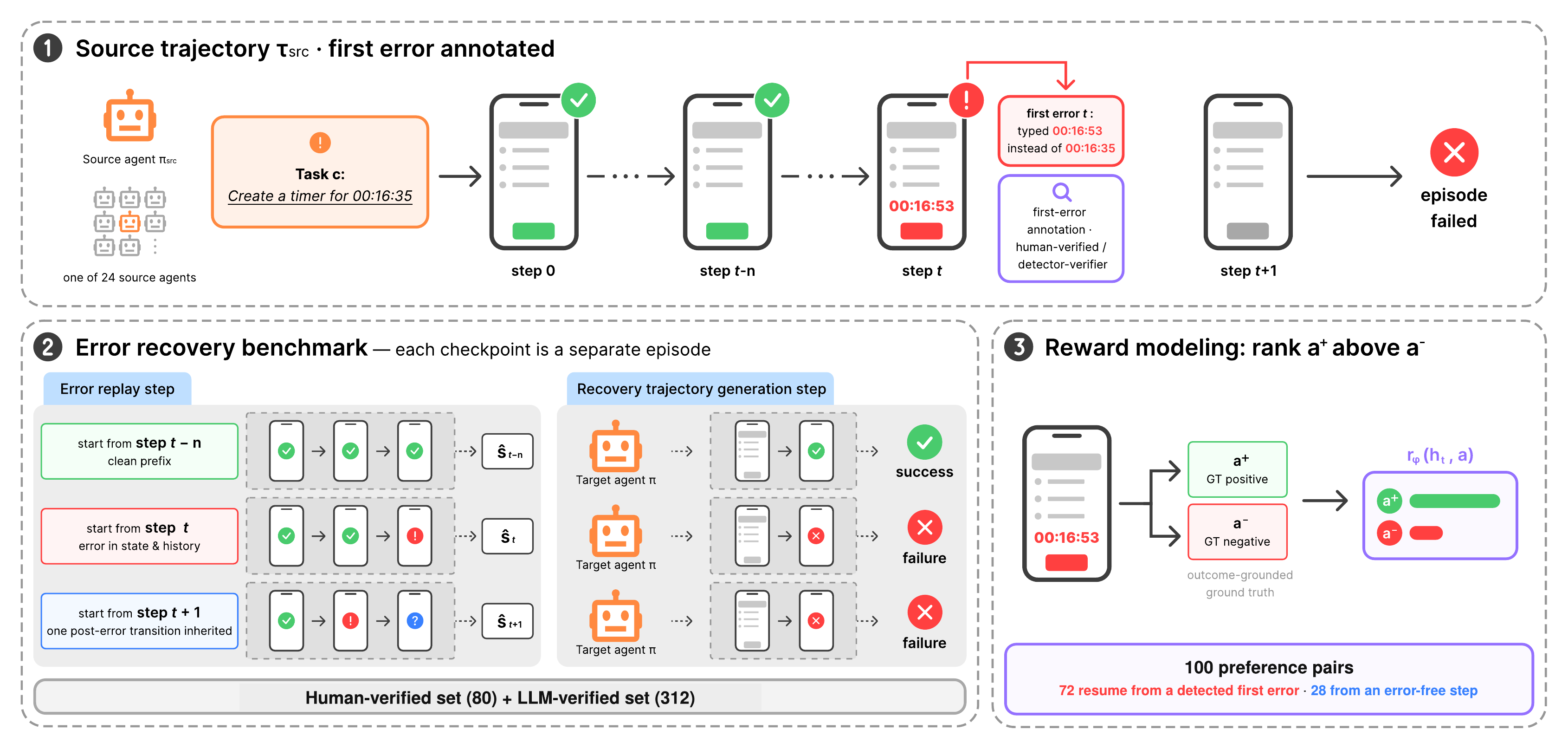}
\caption{\textbf{Overview of AndroidFlux.} A source agent commits a first error at step $t$. The error recovery benchmark replays to a checkpoint around $t$ and resumes a target agent from the restored state. The reward modeling benchmark asks whether a reward model can rank recovery actions.}
\label{fig:overview}
\vspace{-8pt}
\end{figure}

\begin{table*}[t]
\centering
\scriptsize
\setlength{\tabcolsep}{4pt}
\resizebox{\textwidth}{!}{%
\begin{tabular}{llccccclcll}
\toprule
& & & \multicolumn{6}{c}{Error recovery} & & \\
\cmidrule(lr){4-9}
Benchmark & Platform & Live & Eval. & Error src. & Src. agents & Full trans. & $t{+}1$ rec. action & RM eval & Starts at & Scale \\
\midrule
AndroidControl & Mobile  & \xmark & \xmark & --     & --  & \xmark & --        & \xmark & any step & 8,444 steps \\
AndroidWorld   & Mobile  & \cmark & \xmark & --     & --  & \xmark & --        & \xmark & step 0   & 116 tasks \\
AndroidLab     & Mobile  & \cmark & \xmark & --     & --  & \xmark & --        & \xmark & step 0   & 138 tasks \\
MobileWorld    & Mobile  & \cmark & \xmark & --     & --  & \xmark & --        & \xmark & step 0   & 201 tasks \\
OSWorld        & Desktop & \cmark & \xmark & --     & --  & \xmark & --        & \xmark & step 0   & 369 tasks \\
GUI-RobustEval & Desktop & \cmark & \cmark & Policy & 12  & \xmark & Neg. only & \xmark & $t{+}n^{-}$ & 1,216 cases \\
\midrule
\textbf{AndroidFlux (ours)} & \textbf{Mobile} & \cmark & \cmark & \textbf{Policy} & \textbf{21}
  & \cmark & \textbf{Pos. + Neg.} & \cmark & \textbf{$t{-}n$, $t$, $t{+}1^{\pm}$} & \textbf{392 cases, 100 pairs} \\
\bottomrule
\end{tabular}}
\caption{Comparison with other GUI benchmarks. \emph{Full trans.} marks whether all three phases around the first error are covered; \emph{$t{+}1$ rec.\ action} gives which recovery actions are supplied at the post-error state; \emph{RM eval} marks whether the benchmark offers action preference evaluation for reward models.}
\label{tab:benchmark_compare}
\vspace{-8pt}
\end{table*}

\section{Introduction}

Multimodal agents now operate real mobile devices, completing user tasks by reading screens and performing touch and text actions over long horizons \citep{androidworld,androidlab,mobileworld}. Existing benchmarks evaluate these agents almost entirely by task success from a clean start, and the strongest models are nearing saturation on them. Yet, a clear gap still remains between benchmark scores and in-the-wild performance. Part of the reason is that a single pass or fail per episode hides what happens inside it: the intermediate failures behind the outcome and which actions helped or hurt along the way. Because mobile environments change dynamically, a general mobile-use agent must complete tasks robustly even from unexpected states and in unseen environments. This requires knowing when and where an agent makes errors, and whether it can recover from errors and subsequent states. Recovery matters even more in recent agentic systems, which increasingly optimize inference cost by assigning easier tasks to smaller agents \citep{frugalgpt,routellm}. When a smaller agent stalls, these systems often hand over the unfinished task to a stronger model \citep{stepcascade}, which must then recover from mistakes it did not make, in a context and device state it did not produce. However, observing these behaviors through clean-start rollouts is expensive: failure states arise only by chance, each one costs a full episode to reach, and the score reflects the whole episode rather than the specific steps where recovery succeeds or fails.

Although measuring failure recovery is critical for closing the gap to real-world deployment and for building efficient agentic systems, no existing mobile-use benchmark directly evaluates this capability. To address this, we introduce \textbf{AndroidFlux} (Figure~\ref{fig:overview}) to evaluate the failure recovery and recovery-action reward modeling capabilities through controlled replay of policy-induced failures in AndroidWorld. We annotate the first error step $t$ and restore checkpoints at $t{-}n$ (correct prefix), $t$ (error committed), and $t{+}1$ (after a positive or negative recovery action). Target agents resume from matched device states with histories re-rendered in their own formats, allowing us to examine how policies handle the same failure and how recovery depends on the preceding actions. Across human-verified and larger LLM-verified sets, recovery performance only partially correlates with clean-start success, showing that existing benchmarks alone are insufficient for assessing an agent's robustness. We find that explicit Chain-of-Thought (CoT) reasoning also does not consistently improve recovery: general reasoning ability alone does not ensure effective action in a disrupted GUI state. On our reward-modeling benchmark, reward models trained on generic GUI preferences stay near chance, while GUI experts' own action likelihoods rank recovery actions better, and the two are correct on largely different pairs, suggesting they are complementary. Furthermore, we release an offline RL dataset of 8.5k decision contexts that improves both recovery and clean-start success, offering a practical resource for studying failure recovery without expensive online RL training.

\section{Related Work}

\paragraph{Mobile/GUI benchmarks.}
GUI-agent benchmarks differ mainly in whether the agent acts in a live environment and what state each episode begins from. AndroidControl supplies static UI-control data along demonstrations, scorable at any step but without consequences \citep{androidcontrol}. AndroidWorld, AndroidLab, MobileWorld, and OSWorld provide interactive environments with programmatic verifiers, but most episodes start from the home screen and return one terminal outcome \citep{androidworld,androidlab,mobileworld,osworld}. Concurrent with this work, GUI-RobustEval benchmarks post-error recovery on desktop agents by resuming after a policy-induced error \citep{guirobusteval}. AndroidFlux instead targets mobile agents and spans controlled checkpoints before, at, and after the first error, enabling cross-agent takeover from inherited context. It further evaluates outcome-grounded recovery-action preferences, evaluating not only whether an agent recovers but whether a reward model can identify the action that makes recovery more likely. 

\paragraph{Reward modeling.}
Mobile GUI agents are typically trained against sparse terminal success \citep{mobilerl}, while process reward models supply denser signal by scoring intermediate decisions \citep{lightman}, and preference-based reward models are conventionally fit under the Bradley--Terry (BT) interface \citep{bradleyterry,stiennon,instructgpt}. UI-Genie \citep{uigenie} is the closest GUI precedent, training action- and task-level rewards from rule verification and synthetic corruption. Its negatives, however, are perturbations of correct behavior rather than actions a policy took at a state it had actually broken, and the reward model is trained inside a self-improvement loop with a single agent. The sources of errors used for reward modeling have not yet been scaled across policies, and RL with such rewards remains largely unexplored in both online and offline settings.

\section{Problem Formulation}

\paragraph{Preliminaries.}
We model mobile interaction as a finite-horizon POMDP with agent-specific context,
\begin{equation}
\mathcal{M}=(\mathcal{S},\mathcal{O},\mathcal{A},P,\Omega,r,H,\mu_0),
\qquad
h_t^{\pi}=\Phi_{\pi}(c,o_{0:t},y_{0:t-1}),
\end{equation}
where $s_t$ is latent device state, $o_t$ the observation, $a_t$ an AndroidWorld action, $r$
the terminal programmatic verifier, $c$ the task, $y_i=(u_i,a_i)$ a response with optional
reasoning $u_i$, and $\Phi_\pi$ the target agent's context formatting. The screenshot alone does not
provide application databases, activity stacks, or filesystem state, making recovery inherently
history-dependent. Recorded responses are stored schema-neutrally and re-rendered per policy,
so cross-agent contexts do not copy one model's prompt format into another. Further detailed action space of $a_t$ is reported in Appendix~\ref{action_space}.

\paragraph{Replay and resume from the first error.}
For source trajectory $\tau^{\mathrm{src}}$ and checkpoint $k$, replay re-executes the recorded
action prefix under the same task seed, falling back to an emulator snapshot where replay is not
reproducible; $k$ is thus the last replayed step supplied in context, and the target's first new
action occurs at $k+1$. Letting $e_i\in\{0,1\}$ indicate whether $a_i$ fails to advance the task
from $s_i$, the first error is the earliest such step:
\begin{equation}
\hat{s}_k=\mathrm{Replay}(\tau^{\mathrm{src}},k),
\qquad
t=\min\{\,i : e_i=1\,\}.
\end{equation}
Any $y_i$ with $i>t$ is taken at a state already carrying the effect of $a_t$, so it is an
attempt to act from a broken state rather than an independent failure; all such actions are
recovery actions, and $t$ is the only error committed under a correct prefix. We consider only
recoverable first errors.

Resume runs target policy $\pi$ from checkpoint $k$ using the restored state $\hat{s}_k$ and
the reconstructed context $h_k^\pi$, generating a recovery trajectory whose terminal state the
verifier scores:
\begin{equation}
\label{eq:resume}
\tau^{\mathrm{rec}}=\mathrm{Resume}(\pi,\hat{s}_k,h_k^\pi)=(y_{k+1},\dots,y_T),
\qquad
\mathrm{SR}(\pi,k)=\frac{1}{N}\sum_{i=1}^{N}\mathbf{1}\!\left[r(s_{T_i}^{(i)})=1\right].
\end{equation}
The target policy thus recovers from its own context when $\pi=\pi_{\mathrm{src}}$, and from another agent's context when $\pi\neq\pi_{\mathrm{src}}$. Varying $k$ relative to $t$ separates the cost of a committed error from the cost
of inherited foreign history.

\paragraph{Recovery-action reward modeling.}
The error recovery task resumes from the state at $k=t$ and asks which action should be taken next. With $x=(h_t,o_t)$, a reward model assigns a scalar score $r_\phi(x,a)$ to each candidate action, and preferences follow from score ordering. A pair $(a^+,a^-)$ is ranked correctly when $r_\phi(x,a^+)>r_\phi(x,a^-)$, and accuracy is measured over all outcome-labeled pairs. Both actions in a pair share the task, state, and history. 

\section{AndroidFlux: Error Recovery \& Recovery Action Reward Modeling Benchmark}
To measure whether a policy repairs a broken state and whether a reward model can identify the action that repairs it, we introduce \textbf{AndroidFlux}: a failure-recovery benchmark and a recovery-action reward modeling benchmark, both built on the AndroidWorld environment and tasks. Table~\ref{tab:benchmark_compare} overviews them against existing GUI agent benchmarks.

\subsection{Error Recovery Benchmark}
The error recovery benchmark supplies restored resume points drawn from real policy failures. Its human-verified set is fixed at 80 cases from 6 source models, 65 restored by replaying the recorded action prefix and 15 from saved emulator snapshots where exact replay is not reliable. Its LLM-verified set contains 312 replay-based cases over 104 tasks from 21 source models, extending coverage across a wider pool of failure sources.

\subsubsection{Annotation of the First Error}
The first error is the initial error in the trajectory a model executes toward its goal, and the only one unaffected by a preceding error. For the human-verified set, human experts reviewed the trajectory step by step and labeled the earliest error. For the LLM-verified set, a two-stage detector--verifier framework implemented with Gemini Flash 3.0~\citep{gemini3flash} proposes candidate error steps and re-checks each against the resulting UI state, taking the earliest surviving step as the first error. Because a false label relocates the state at which every downstream recovery experiment begins, the framework is tuned toward precision at the expense of recall, reaching 90.8\% step-level precision on the annotated set and so identifying policy-induced first errors reliably. Each error is classified to one of three categories---Stagnation, Direction, and Execution. Further details are reported in Appendix~\ref{error_category} and~\ref{detection_framework}.
\vspace{-8pt}

\subsection{Recovery Action Reward Modeling Benchmark}
The AndroidFlux recovery-action reward modeling benchmark fixes the resume state and pairs a positive and a negative next action, and a reward model is expected to rank the pair by which action is more likely to lead to recovery. It contains 100 recovery states over 26 AndroidWorld task types: 72 resume from a detected first error and 28 from an error-free step. We generate source trajectories from ten agents, annotate their first errors with the detector--verifier pipeline, and resume five recovery models from the restored states to obtain candidate recovery actions. Each action candidate is replayed and a panel of twelve agents spanning a wide range of recovery ability rolls out from the resulting state. Preference thus follows from consequence rather than rule or judge: at each state we pair the two candidates whose panel success counts differ most, keeping the pair only when the margin is at least two. The resulting 100 pairs share task, state, screenshots, and history other than the recovery action. Further details are in Appendix~\ref{rm_apd}.

\subsection{Dataset: Offline RL Training Dataset for Error Recovery}

The RL dataset contains 8,499 policy-induced contexts built from source trajectories collected before replay filtering, sampled with seed 42 from four slices: $t{-}n$ (2,097), $t{-}1$ and $t$ (2,080 each), and successful contexts with verified environment success (2,242). Each is annotated with whether its goal is already satisfied, making termination the correct next action. A checkpoint is the last action in the history, so $t{-}1$ prompts the erroneous action and $t$ prompts a recovery at $t{+}1$; no history extends past the first error. Construction and the training details are given in Appendix~\ref{app:offline_rl_data} for details.

\section{Main Results}
\label{main_results}
All experiments run in the containerized AndroidWorld environment released with the benchmark, which we customize for reproducibility and stability. Every evaluation in this paper is split across container shards running on H100, A40, L40S, and RTX 4090 GPUs.

\subsection{Failure Recovery Across Agent Architectures}


Table~\ref{tab:modelclass} shows a clear progression in recovery capability across agent architectures. General-purpose VLMs can recover from inherited failures, but domain-specialized GUI agents are generally more robust around the first-error checkpoint. On the human-verified set, GUI-Owl-1.5-8B achieves $40.0$ at $t_{\text{all}}$, compared with $28.8$, $30.0$, and $27.5$ for Qwen3-VL-8B, Qwen3-VL-32B, and Gemini Flash 3.0. The same pattern holds from $t{-}1$ to $t$: most models are affected by the committed error, but stronger expert policies retain more of their pre-error performance. This suggests that training for interactive GUI control provides robustness to off-distribution states beyond clean-start capability. Agentic frameworks strengthen this effect through inference-time state reassessment and replanning. Mobile-Agent-v3-7B reaches $35.4$ at $t_{\mathrm{rep}}$, compared with $24.6$ for GUI-Owl-7B, with the gap widening on the LLM-verified set. MobileRun provides the strongest example: using Gemini Flash 3.0 within the framework raises recovery from $27.5$ to $67.5$ at $t_{\text{all}}$ on the human-verified set and from $39.7$ to $67.0$ at $t$ on the LLM-verified set. It also remains strong after either injected recovery action. These results show that recovery depends not only on the underlying model, but also on inference-time scaling mechanisms for reflection, correction, and replanning.

Table~\ref{tab:instreason} and Figure~\ref{fig:reasoning} compare matched instruct and explicit CoT reasoning variants within each model family, with Gemini Flash 3.0 providing a same-model thinking on/off control. Explicit CoT does not uniformly improve recovery: Qwen3-VL-32B and Gemini Flash 3.0 perform worse at $t$ with reasoning enabled, whereas GUI-Owl-1.5-32B and GPT-5.6 Terra improve substantially. This suggests that reasoning acquired through general-domain RL does not necessarily transfer to interactive GUI recovery. In contrast, GUI-Owl and GPT-5.6 Terra apply reasoning-oriented RL in the target GUI domain, and their reasoning variants consistently improve recovery, indicating that domain-aligned RL enhances error recovery performance.

\definecolor{clsbg}{RGB}{229,236,247}   
\definecolor{revred}{RGB}{190,30,35}
\begin{table}[!t]
\setlength{\abovecaptionskip}{2pt}
\setlength{\belowcaptionskip}{1pt}
\centering\scriptsize\setlength{\tabcolsep}{3pt}
\resizebox{\ifdim\width>\textwidth \textwidth\else\width\fi}{!}{%
\begin{tabular}{llcccccccc ccccc}
\toprule
 & & \multicolumn{8}{c}{Human-verified} & \multicolumn{5}{c}{LLM-verified} \\
\cmidrule(lr){3-10} \cmidrule(lr){11-15}
Type & Model & Scratch & $t{-}1_{\text{rep}}$ & $t_{\text{rep}}$ & $t_{\text{snap}}$ & $t_{\text{all}}$ & $t{+}1^{+}$ & $t{+}1^{-}$ & $t{+}1_{\text{all}}$
             & Scratch & $t{-}1$ & $t$ & $t{+}1^{+}$ & $t{+}1^{-}$ \\
\midrule
\multirow{10}{*}{General}
& Qwen3-VL-8B-Instruct    & 41.25 & 33.85 & 24.62 & 46.67 & 28.75 & 36.54 & 26.92 & 31.73 & 49.04 & 36.86 & 29.49 & 48.56 & 43.27 \\
 & Qwen3-VL-32B-Instruct   & 36.25 & 36.92 & 26.15 & 46.67 & 30.00 & 38.46 & 30.77 & 34.62 & 45.19 & 42.63 & 33.97 & 50.96 & 48.56 \\
 & Qwen3.8-27B  & 45.00 & 30.77 & 27.69 & 53.33 & 32.50 & 40.38 & 30.77 & 35.58 & 50.00 & 47.12 & 37.82 & 57.21 & 50.00 \\
 & Gemma-4-31B-IT (M3A)    & 37.50 & 40.00 & 36.92 & 66.67 & 42.50 & 48.08 & 36.54 & 42.31 & 43.27 & 51.60 & 44.55 & 62.50 & 61.06 \\
 & Gemini Flash 3.0$^{\dagger}$        & 40.00 & 29.23 & 24.62 & 40.00 & 27.50 & 42.31 & 34.62 & 38.46 & 46.15 & 41.35 & 39.74 & 49.52 & 48.56 \\
 & Gemini Flash 3.8$^{\dagger}$        & 51.25 & 52.31 & 50.77 & 66.67 & 53.75 & 50.00 & 53.85 & 51.92 & 58.65 & 58.33 & 61.54 & 74.52 & 69.23 \\
 & Gemini 3.1 Pro$^{\dagger}$          & 55.00 & 55.38 & 47.69 & 60.00 & 50.00 & 63.46 & 51.92 & 57.69 & 61.54 & 62.18 & 58.33 & 78.85 & 65.87 \\
 & GPT-5.6 Luna$^{\dagger}$          & 27.50 & 33.85 & 27.69 & 60.00 & 33.75 & 28.85 & 21.15 & 25.00 & 33.65 & 35.26 & 31.73 & 41.35 & 40.38 \\
 & GPT-5.6 Terra$^{\dagger}$         & 38.75 & 40.00 & 43.08 & 53.33    & 45.00    & 46.15 & 38.46 & 42.31 & 53.85 & 46.15 & 42.63 & 60.10 & 59.62 \\
\midrule
\multirow{4}{*}{Expert}
 & MAI-UI-8B               & 37.50 & 40.00 & 30.77 & 53.33 & 35.00 & 38.46 & 38.46 & 38.46 & 46.15 & 41.35 & 31.73 & 50.96 & 48.56 \\
 & GUI-Owl-7B              & 46.25 & 41.54 & 24.62 & 53.33 & 30.00 & 36.54 & 34.62 & 35.58 & 56.73 & 40.38 & 27.24 & 45.19 & 41.83 \\
 & GUI-Owl-1.5-8B \citep{guiowl}
                           & 56.25 & 47.69 & 36.92 & 53.33 & 40.00 & 53.85 & 46.15 & 50.00 & 62.50 & 49.68 & 44.23 & 62.98 & 58.65 \\
 & Gemini Flash 3.5 Computer Use$^{\dagger}$ \citep{gemini35computeruse}
                           & 38.75 & 30.77 & 30.77 & 53.33 & 35.00 & 61.54 & 63.46 & 62.50 & 40.38 & 57.69 & 40.06 & 74.04 & 70.19 \\
\midrule
\multirow{3}{*}{\shortstack[l]{Agentic\\framework}}
 & Mobile-Agent-v3-7B      & 58.75 & 44.62 & 35.38 & 53.33 & 38.75 & 51.92 & 40.38 & 46.15 & 63.46 & 49.68 & 45.19 & 61.54 & 56.73 \\
 & MobileRun \citep{mobilerun}
                           & 78.75 & 63.08 & 64.62 & 80.00 & 67.50 & 82.69 & 75.00 & 78.85 & 79.81 & 69.87 & 66.99 & 75.96 & 82.21 \\
 & FinalRun \citep{finalrun}
                           & 66.25 & 38.46 & 40.00 & 60.00 & 43.75 & 42.31 & 34.62 & 38.46 & 68.27 & 43.27 & 39.10 & 59.13 & 54.33 \\
\bottomrule
\end{tabular}}
\caption{General-purpose VLMs, GUI experts, and agentic frameworks on error recovery on both human-verified set and LLM-verified set.
The columns follow the recovery checkpoints defined in
Table~\ref{tab:hvcomparison}.
All agentic frameworks are driven by Gemini Flash 3.0 and all proprietary models$^{\dagger}$ are evaluated with reasoning enabled and Qwen3-VL AndroidWorld prompt.
}\label{tab:modelclass}
\vspace{-4pt}
\end{table}

\definecolor{guibg}{RGB}{233,244,233}
\definecolor{guigrn}{RGB}{26,102,38}
\newsavebox{\instreasonbox}

\begin{figure}[t]
\setlength{\abovecaptionskip}{3pt}\setlength{\belowcaptionskip}{0pt}
\setlength{\aboverulesep}{0.15ex}\setlength{\belowrulesep}{0.15ex}
%
\begin{minipage}[t]{0.52\linewidth}
\vspace{0pt}\centering\scriptsize\setlength{\tabcolsep}{2.5pt}
\renewcommand{\arraystretch}{0.90}
\global\setbox\instreasonbox=\hbox{\resizebox{\ifdim\width>\linewidth \linewidth\else\width\fi}{!}{%
\begin{tabular}{lcccccc}
\toprule
& \multicolumn{3}{c}{Human-verified $t$} & \multicolumn{3}{c}{LLM-verified $t$} \\
\cmidrule(lr){2-4} \cmidrule(lr){5-7}
Model & Inst. & Reas. & $\Delta$ & Inst. & Reas. & $\Delta$ \\
\midrule
\multicolumn{7}{l}{\textit{Without GUI Domain Specific Reasoning RL}}\\
Qwen3-VL-32B     & 30.00 & 17.50 & \textcolor{revred}{$-12.50$} & 33.97 & 21.15 & \textcolor{revred}{$-12.82$} \\
Gemini Flash 3.0 & 41.25 & 27.50 & \textcolor{revred}{$-13.75$} & 46.79 & 39.74 & \textcolor{revred}{$-7.05$} \\
Qwen3-VL-8B      & 28.75 & 16.25 & \textcolor{revred}{$-12.50$} & 29.49 & 27.88 & \textcolor{revred}{$-1.60$} \\
Gemma-4-31B-IT (M3A)   & 42.50 & 41.25 & \textcolor{revred}{$-1.25$}  & 44.55 & 47.44 & $+2.88$ \\
\midrule
\multicolumn{7}{l}{\textit{With GUI Domain Specific Reasoning RL}}\\
\rowcolor{guibg} GUI-Owl-1.5-32B & 32.50 & 37.50 & \textcolor{guigrn}{$\mathbf{+5.00}$} & 36.22 & 44.55 & \textcolor{guigrn}{$\mathbf{+8.33}$} \\
\rowcolor{guibg} GPT-5.6 Terra & 27.50 & 45.00 & \textcolor{guigrn}{$\mathbf{+17.50}$} & 32.05 & 42.63 & \textcolor{guigrn}{$\mathbf{+10.58}$} \\
\bottomrule
\end{tabular}}}\usebox{\instreasonbox}
\captionof{table}{Instruct vs.\ reasoning modes at the first-error checkpoint $t$,
for models without and with GUI domain specific reasoning RL.}
\label{tab:instreason}
\end{minipage}%
\hfill
\begin{minipage}[t]{0.46\linewidth}
\vspace{0pt}\centering
\includegraphics[height=\dimexpr\ht\instreasonbox+\dp\instreasonbox\relax,
                 width=\linewidth,keepaspectratio]{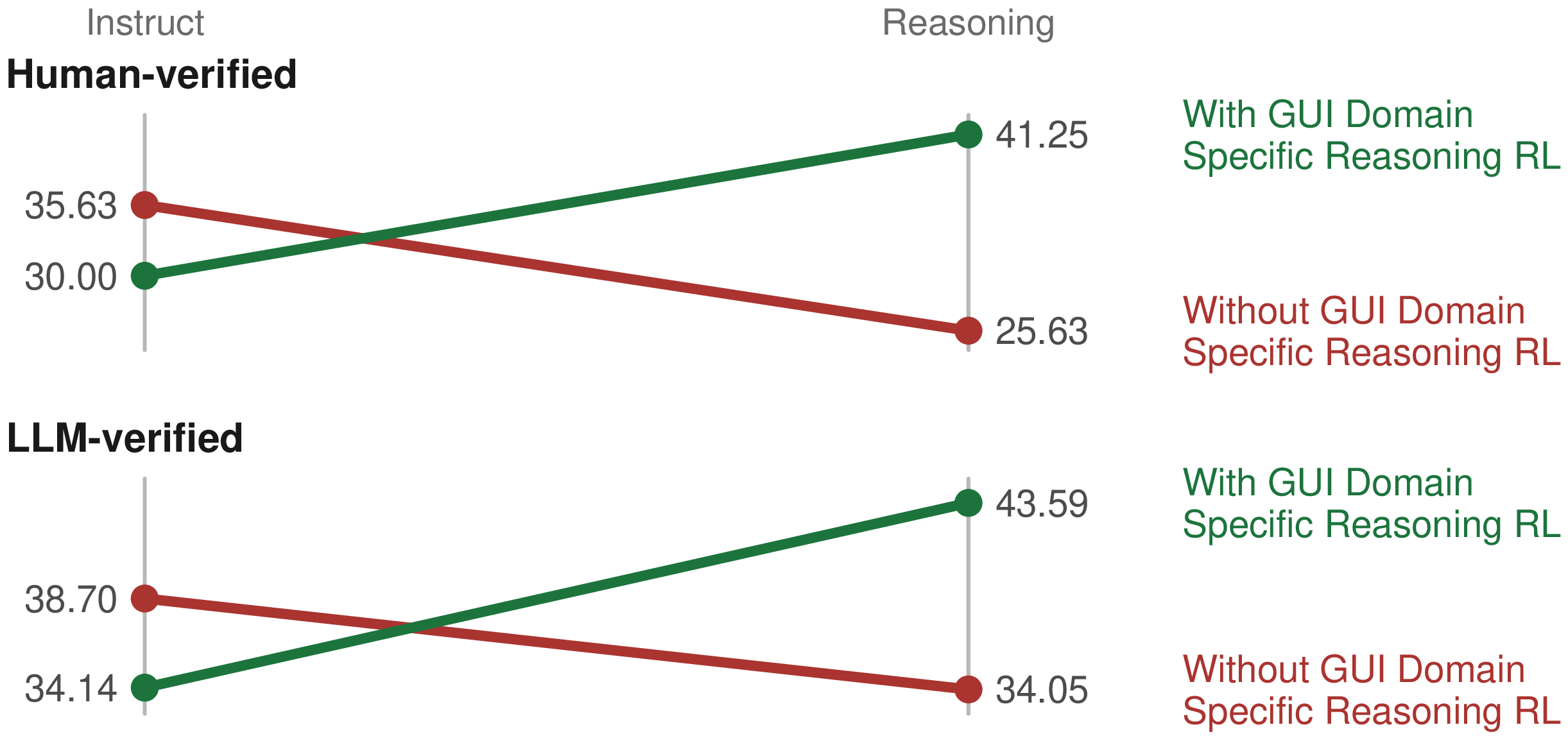}
\caption{Average of Table~\ref{tab:instreason}: models with the domain specific reasoning RL gain, models without lose.}
\label{fig:reasoning}
\end{minipage}
\vspace{-4pt}
\end{figure}

\subsection{Rewinding vs.\ Injection by Failure Timing}
\label{app:reweind_injection}

 An agent's mistake can be remedied in two ways. \emph{Rewinding} returns the agent to the state just before the error and lets it attempt the step again; \emph{injection} leaves the error in place but hands the agent one correct recovery action and lets it continue from there. In our checkpoints, the value of rewinding is the gap between $t{-}1$ and $t$, and the value of injection is the gap between $t{+}1^{+}$ and $t{+}1^{-}$. Which remedy helps
more depends on how far into the task the failure occurs.Splitting the 312 episodes by how many steps the agent had taken before the error, rewinding is most valuable for early failures ($+8.76$pp) and loses most of its benefit for late ones ($+3.41$pp), whereas injection is nearly useless early on and becomes the more helpful remedy late ($+5.86$pp; Table~\ref{tab:prefix}, Figure~\ref{fig:prefix}). The gap between the two grows by $12.22$pp, with a 95\% confidence interval of $[+4.1,+20.8]$. Because every model attempts every episode, results within an episode are
correlated, so the interval resamples episodes rather than individual model runs. The direction of the effect is robust: it holds for 17 of the 18 models individually, across all task-difficulty quartiles, across all trajectory-length terciles, and in both halves of a random episode split. The length-tercile check is the important one, since longer tasks also have longer prefixes; the effect appears within each tercile, so it is not simply a proxy for task length. Within these finer slices the gap is always positive (from $+3.29$pp to $+21.03$pp) but not always individually significant, since each slice is small; we read them as evidence of a consistent direction rather than of significance in every subgroup. Bucket boundaries were fixed in advance rather than searched. Intuitively, an early mistake costs little context, so a second attempt is cheap and effective; a late mistake sits on top of a long history, and what the agent most needs is not another chance to reason but the one action that gets it back on track.

\definecolor{conbg}{RGB}{237,241,247}
\newsavebox{\prefixbox}
\begin{figure}[t]
\centering
\setlength{\abovecaptionskip}{2pt}\setlength{\belowcaptionskip}{0pt}
\setlength{\aboverulesep}{0.25ex}\setlength{\belowrulesep}{0.25ex}
\begin{minipage}[t]{0.46\linewidth}
\vspace{0pt}\centering\footnotesize
\setlength{\tabcolsep}{4pt}\renewcommand{\arraystretch}{1.05}
\global\setbox\prefixbox=\hbox{\resizebox{\ifdim\width>\linewidth \linewidth\else\width\fi}{!}{%
\begin{tabular}{@{}lrrrr@{}}
\toprule
Steps before & & \multicolumn{3}{c}{Gain (pp)} \\
\cmidrule(l){3-5}
the error & $n$ & Rewind & Inject. & $\Delta$ \\
\midrule
$\le 3$   & 116 & $+8.76$ & $-1.01$ & $-9.77$ \\
$4$--$6$  & 100 & $+7.67$ & $+7.55$ & $-0.11$ \\
$\ge 7$   &  96 & $+3.41$ & $+5.86$ & $+2.44$ \\
\midrule
\rowcolor{conbg}\multicolumn{4}{@{}l}{Latest $-$ earliest} & $\mathbf{+12.22}$ \\
\bottomrule
\end{tabular}}}\usebox{\prefixbox}
\captionof{table}{Gain from rewinding ($t{-}1$ vs.\ $t$) and injection
($t{+}1^{+}$ vs.\ $t{+}1^{-}$) by steps taken before the error.}
\label{tab:prefix}
\end{minipage}%
\hfill
\begin{minipage}[t]{0.51\linewidth}
\vspace{0pt}\centering
\includegraphics[height=\dimexpr\ht\prefixbox+\dp\prefixbox\relax,
                 width=\linewidth,keepaspectratio]{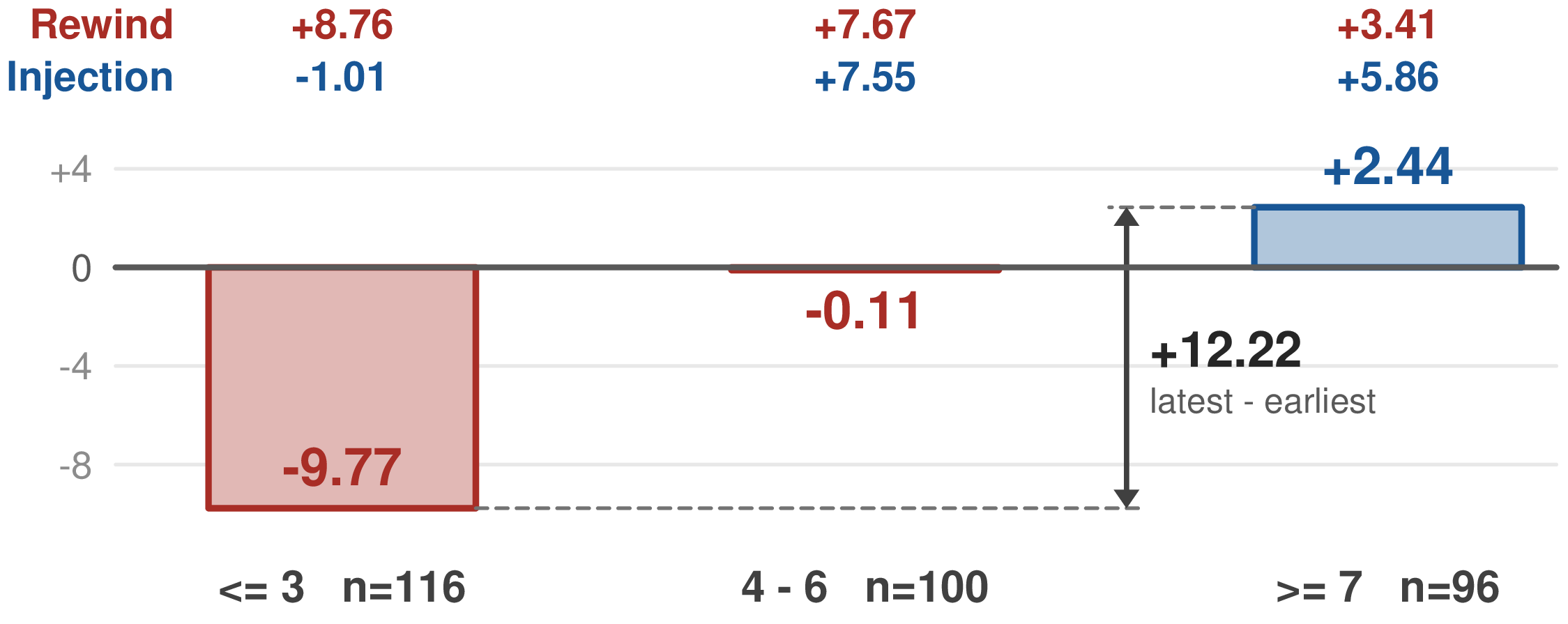}
\caption{$\Delta$ rises monotonically with failure step: rewinding beats
injection on the earliest failures, injection wins on the latest.}
\label{fig:prefix}
\end{minipage}
\end{figure}

\begin{table}[!tbp]
\centering\scriptsize\setlength{\tabcolsep}{3pt}
\resizebox{\linewidth}{!}{%
\begin{tabular}{l*{11}{c}}
\toprule
& \multicolumn{5}{c}{Single agent} & \multicolumn{4}{c}{UI-Genie RM} & \multicolumn{2}{c}{Oracle} \\
\cmidrule(lr){2-6}\cmidrule(lr){7-10}\cmidrule(lr){11-12}
& MAI-UI & GUI-Owl & GUI-Owl & Qwen3-VL & Qwen3-VL & BT & BT & SFT-128K & SFT-128K & 5 agents & 5 agents \\
& 8B & 1.5-8B & 7B & 8B & 2B & 128K & 256K & Hard & Soft & & + BT-256K \\
\midrule
UI-Genie                 & 0.736 & 0.744 & 0.705 & 0.747 & 0.679 & 0.904 & 0.915 & 0.681 & 0.878 & 0.928 & 0.989 \\
AndroidFlux        & 0.600 & 0.590 & 0.580 & 0.520 & 0.520 & 0.490 & 0.520 & 0.160 & 0.450 & 0.820 & 0.870 \\
\quad $\llcorner$ Error & 0.597 & 0.569 & 0.569 & 0.500 & 0.514 & 0.444 & 0.486 & 0.097 & 0.403 & 0.819 & 0.875 \\
\quad $\llcorner$ Clean & 0.607 & 0.643 & 0.607 & 0.571 & 0.536 & 0.607 & 0.607 & 0.321 & 0.571 & 0.821 & 0.857 \\
\bottomrule
\end{tabular}}
\caption{Pairwise accuracy on UI-Genie (1,000 pairs) and AndroidFlux (100 pairs), split into error-path and clean-path pairs. \emph{Single agent} ranks by likelihood, \emph{UI-Genie RM} is trained on UI-Genie, and \emph{Oracle} is correct if any listed scorer is. Details are in Appendix~\ref{rm_apd}.}
\label{tab:rm_main}
\vspace{-16pt}
\end{table}

\subsection{Recovery-Action Reward Modeling}

Table~\ref{tab:rm_main} compares training-free scorers, which rank actions by the log-probabilities of expert agents, with trained reward models. On UI-Genie, BT outperforms SFT trained on the same 128K samples from 64K pairs ($0.904$ vs.\ $0.878$) with LoRA~\citep{hu2022lora} on Qwen3-VL-8B. The hard SFT only checks whether the predicted correct/incorrect token matches the label, so it cannot order two actions that receive the same token. We therefore also compare the log-probability each action assigns to that token (Soft). Even so, SFT still under-performs BT, which suggests that SFT learns absolute correctness well, while relative preference is better learned through the pairwise BT loss. Scale also matters. Scaling the data from 128K to 256K adds 1.1 points on UI-Genie and 3 points on AndroidFlux on UI-Genie RM BT, so AndroidFlux still separates reward models where UI-Genie is close to saturation. Still, both reward models stay near chance on AndroidFlux. This indicates that UI-Genie’s preference data, built by injecting errors into clean trajectories, does not provide enough signal to learn recovery. The GUI experts, in contrast, already hold these preferences. Without any reward-model training, their likelihoods reach $0.71$--$0.74$ on UI-Genie and $0.58$--$0.60$ on AndroidFlux. Their advantage over BT-256K comes almost entirely from error-path pairs, where each expert ranks 6--8 more of the 72 pairs correctly. Finally, these sources are complementary. The best single scorer ranks 60 of the 100 pairs correctly, but at least one of the five agents ranks 82 correctly, and adding BT-256K raises this to 87, because agent likelihoods and the UI-Genie reward model fail on different pairs. Combining the agents’ preferences with the more general preferences of a UI-Genie reward model could cover most of the preferences in AndroidFlux. Further details are in Appendix~\ref{rm_apd}.

\subsection{Offline Reinforcement Learning}

We further examine whether training on the AndroidFlux offline RL dataset improves error recovery and clean-start success. Building on prior work that optimizes VLM and LLM based agents with offline RL~\citep{qian2025toolrlrewardtoollearning, chen2025tgpo} and online RL~\citep{feng2025retoolreinforcementlearningstrategic, yang2026openwebrl}, we finetune Qwen3-VL-4B with LoRA using offline GRPO with Group Preference Reward Shaping (GPRS)~\citep{zhu2025gprs}, without executing sampled actions in the environment during training. We use UI-Genie BT-128K as the reward model and fix the training budget at 2.4K samples. We compare UI-Genie reward modeling data with AndroidFlux offline RL data collected at the first-error checkpoint $t$, $t{-}1$, and mixtures of these sources. Further details are provided in Appendix~\ref{app:offline_rl_details}.

\newsavebox{\rlbox}
\begin{figure}[t]
\centering
\setlength{\abovecaptionskip}{2pt}\setlength{\belowcaptionskip}{0pt}
\setlength{\aboverulesep}{0.25ex}\setlength{\belowrulesep}{0.25ex}
\begin{minipage}[t]{0.46\linewidth}
\vspace{0pt}\centering\footnotesize
\setlength{\tabcolsep}{4pt}\renewcommand{\arraystretch}{1.05}
\global\setbox\rlbox=\hbox{\resizebox{\ifdim\width>\linewidth \linewidth\else\width\fi}{!}{%
\begin{tabular}{@{}lccc@{}}
\toprule
Training data & Clean & Error-$t$ & Mean \\
\midrule
Baseline (Qwen3-VL 4B)                & 20.0 & 21.5 & 20.8 \\
UI 100\%                              & 23.1 & 16.9 & 20.0 \\
$t$ 100\%                             & 18.5 & \textbf{30.8} & 24.6 \\
UI 75\% / $t$ 25\%                    & 23.1 & 24.6 & 23.8 \\
UI 50\% / $t$ 25\% / $t{-}1$ 25\%     & \textbf{29.2} & 21.5 & 25.4 \\
UI 50\% / $t$ 35\% / $t{-}1$ 15\%     & 26.2 & 26.2 & 26.2 \\
UI 40\% / $t$ 35\% / $t{-}1$ 25\%     & 27.7 & 26.2 & \textbf{26.9} \\
\bottomrule
\end{tabular}}}\usebox{\rlbox}
\captionof{table}{\textbf{Offline RL Experiments.} Success rate (\%) on 65 HV tasks. Training details are provided in Appendix~\ref{app:offline_rl_details}.}
\label{tab:offline_rl}
\end{minipage}%
\hfill
\begin{minipage}[t]{0.51\linewidth}
\vspace{0pt}\centering
\includegraphics[height=\dimexpr\ht\rlbox+\dp\rlbox\relax,
                width=\linewidth,keepaspectratio]{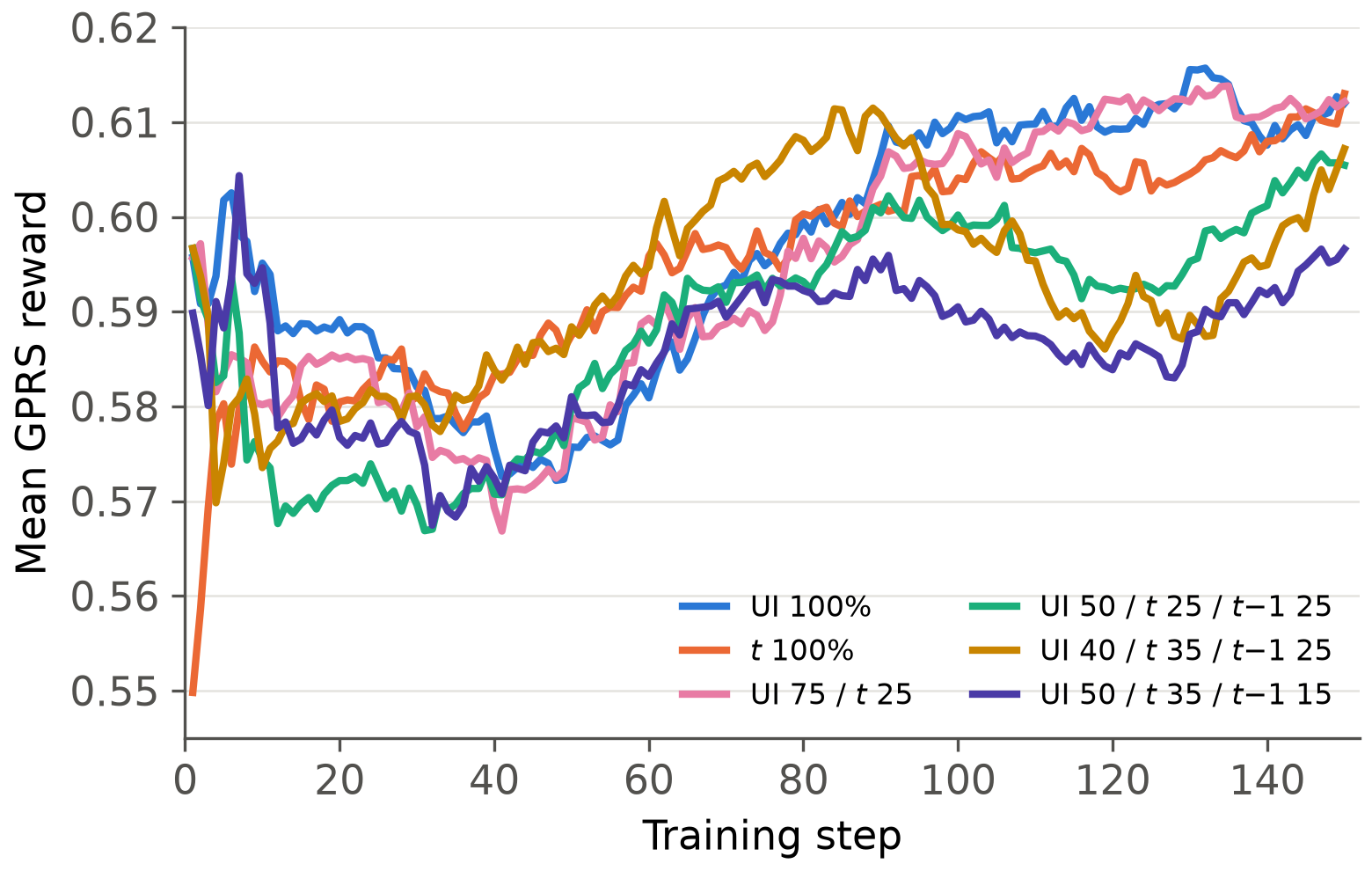}
\caption{Mean GPRS reward per training step. Rewards rise similarly across mixtures despite different downstream success.}
\label{fig:rl_reward}
\end{minipage}
\vspace{-12pt}
\end{figure}

Table~\ref{tab:offline_rl} shows that the source of offline training contexts substantially affects downstream behavior. Training on the UI-Genie reward modeling dataset alone provides little overall benefit and reduces recovery from first-error states. In contrast, every run with our curated offline RL dataset outperforms the UI-Genie 100\% run in mean success. Training exclusively at the first-error checkpoint produces the strongest recovery but degrades clean-start performance, indicating that concentrating only on failure states can over-specialize the policy.

Combining the conventional reward modeling dataset with our policy-induced, failure-centered contexts largely removes this trade-off. In particular, the UI 40\% / $t$ 35\% / $t{-}1$ 25\% mixture improves both clean-start execution and error recovery over the UI-Genie 100\% run, while also outperforming the base policy under both evaluation conditions. Moreover, most mixtures improve clean performance despite using fewer UI-Genie samples, suggesting that states surrounding first errors provide useful supervision beyond recovery itself. These results suggest that our offline RL dataset provides a more persistent learning signal in both clean runs and failure recovery.

Despite the performance differences on both the UI-Genie and AndroidFlux reward modeling evaluation sets, Figure~\ref{fig:rl_reward} indicates that optimization remains stable across all data mixtures. The mean GPRS reward rises similarly in every run, but it is not a direct measure of policy quality, since the statistic is inflated by duplicate sampled actions within each group. Indeed, the UI-Genie 100\% run reaches a comparable final GPRS reward despite the weakest downstream performance. The underlying reward-model scores show a clearer difference: UI-Genie-only training plateaus and then declines later in training, whereas most runs with diverse contexts around the error step keep improving through the final training window, ending with a larger net gain.

Overall, these results indicate that our policy-induced failure contexts provide an informative training signal in offline RL for generalized GUI agents without computationally heavy online environments.

\section{Limitations and Conclusion}
We introduced AndroidFlux, a benchmark that evaluates mobile-use agents on recovering from policy-induced first errors and reward models on ranking recovery actions. However, it covers only AndroidWorld and first recoverable errors that can be reconstructed by replay or snapshots, so chains of interacting errors and other GUI environments remain open. The LLM-verified set also inherits noise and source-model imbalance from the detector--verifier pipeline, and recovery success remains sensitive to Android nondeterminism. Even so, our results show that recovering from a committed error is a capability distinct from clean-start success, with domain-specific RL and agentic frameworks far more robust than general-purpose or imitation-trained models. Our reward modeling benchmark demonstrates that reward models trained on conventional GUI preferences stay near chance on recovery states, while the likelihoods of GUI expert agents rank recovery actions more accurately. Furthermore, we find that our curated offline RL dataset improves both clean-start and recovery success under a fixed training budget, showing that policy-induced failure states are effective training signal. Together, our contributions and findings establish failure recovery and recovery-aware reward modeling as key evaluation targets for building robust GUI agents.

\clearpage
\appendix

{\large\bfseries Appendix}

\section{AndroidWorld \& AndroidFlux Action Space}
\label{action_space}
AndroidFlux inherits the action space of AndroidWorld, summarized in Table~\ref{tab:action_space}. Every action carries an \texttt{action\_type} together with typed arguments, and \texttt{status} and \texttt{answer} terminate the episode. Fixing one grammar is what makes cross-agent replay well defined---candidate actions produced by different agents become comparable strings over a single action language, so a recorded action can be re-executed by the replay operator and a recovery action produced by one model can be scored against one produced by another.

\begin{table}[!h]
\centering\scriptsize\setlength{\tabcolsep}{4pt}
\caption{The action space of AndroidWorld and AndroidFlux. Actions with $\dagger$ terminate the episode.}
\label{tab:action_space}
\begin{tabular*}{\linewidth}{@{\extracolsep{\fill}}lll@{}}
\toprule
Action type & Typed arguments & Effect \\
\midrule
\texttt{click} & \texttt{index} or \texttt{(x,y)} & Tap a UI element or screen coordinate \\
\texttt{double\_tap} & \texttt{index} or \texttt{(x,y)} & Double tap \\
\texttt{long\_press} & \texttt{index} or \texttt{(x,y)} & Press and hold \\
\texttt{scroll} & \texttt{direction} & Scroll the focused container \\
\texttt{swipe} & \texttt{direction} & Swipe across the screen \\
\texttt{input\_text} & \texttt{text}, \texttt{index} & Enter text into a field \\
\texttt{keyboard\_enter} & -- & Submit the soft keyboard \\
\texttt{open\_app} & \texttt{app\_name} & Launch an application \\
\texttt{navigate\_home} & -- & Return to the launcher \\
\texttt{navigate\_back} & -- & Back navigation \\
\texttt{wait} & -- & Yield one step for asynchronous state \\
\texttt{answer}$^{\dagger}$ & \texttt{text} & Return a textual answer \\
\texttt{status}$^{\dagger}$ & \texttt{goal\_status} & Declare the episode complete or infeasible \\
\bottomrule
\end{tabular*}
\end{table}

\section{AndroidWorld}
\label{app:models}
\begin{table}[h]
\centering\scriptsize\setlength{\tabcolsep}{3pt}
\caption{Completed AndroidWorld clean-start measurements (116 tasks).}
\label{tab:appendix_aw}
\resizebox{\textwidth}{!}{%
\begin{tabular}{llrrrrr}
\toprule
Model & Type & Reported SR (\%) & Emulator Correct & Emulator SR (\%) & Docker Correct & Docker SR (\%) \\
\midrule
Qwen3-VL-2B-Instruct & General & 36.4 & 38/116 & 32.76 & 33/116 & 28.45 \\
Qwen3-VL-4B-Instruct & General & 45.3 & 55/116 & 47.41 & 51/116 & 43.97 \\
Qwen3-VL-8B-Instruct & General & 47.6 & 56/116 & 48.28 & 63/116 & 54.31 \\
Qwen3-VL-30B-A3B-Instruct & General & 54.3 & 56/116 & 48.28 & 27/116 & 23.28 \\
Qwen3-VL-32B-Instruct & General & 57.3 & N/R & N/R & 59/116 & 50.86 \\
Qwen3.6-27B  & General & 70.3 & N/R & N/R & 32/116 & 27.59 \\
Qwen3.8-27B  & General & 81.9 & N/R & N/R & 63/116 & 54.31 \\
Gemma-4-31B-IT (M3A) & General & N/P & N/R & N/R & 56/116 & 48.28 \\
Gemini Flash 3.0 (Qwen3-VL prompt) & General & N/P & 48/116 & 41.38 & 57/116 & 49.14 \\
Gemini Flash 3.5 (Qwen3-VL prompt) & General & N/P & N/R & N/R & 76/116 & 65.52 \\
Gemini Flash 3.8 & General & N/P & N/R & N/R & 72/116 & 62.07 \\
Gemini Pro 3.1 & General & N/P & N/R & N/R & 76/116 & 65.52 \\
\midrule
GUI-Owl-7B & Expert & 66.4 & 70/116 & 60.34 & 71/116 & 61.21 \\
GUI-Owl-32B & Expert & N/P & 72/116 & 62.07 & 75/116 & 64.66 \\
GUI-Owl-1.5-2B & Expert & 67.9 & 77/116 & 66.38 & 78/116 & 67.24 \\
GUI-Owl-1.5-4B & Expert & 69.8 & 82/116 & 70.69 & 74/116 & 63.79 \\
GUI-Owl-1.5-8B & Expert & 69.0 & 81/116 & 69.83 & 77/116 & 66.38 \\
GUI-Owl-1.5-32B & Expert & 69.4 & 88/116 & 75.86 & 77/116 & 66.38 \\
MAI-UI-2B & Expert & 49.1 & 51/116 & 43.97 & 44/116 & 37.93 \\
MAI-UI-8B & Expert & 70.7 & 71/116 & 61.21 & 60/116 & 51.72 \\
UI-TARS-1.0-7B-SFT & Expert & 33.0 & 29/116 & 25.00 & 31/116 & 26.72 \\
UI-TARS-1.5-7B & Expert & N/P & 28/116 & 24.14 & 25/116 & 21.55 \\
Gemini Flash 3.5 Computer Use & Expert & N/P & 61/116 & 52.59 & 45/116 & 38.79 \\
\midrule
Mobile-Agent-v3-7B & Agent & N/P & N/R & N/R & 78/116 & 67.24 \\
Mobile-Agent-v3-32B & Agent & 73.3 & 79/116 & 68.10 & 76/116 & 65.52 \\
MobileRun (Gemini Flash 3.0) & Agent & N/P & N/R & N/R & 95/116 & 81.90 \\
FinalRun \citep{finalrun} & Agent & 97.4 & N/R & N/R & 83/116 & 71.55 \\
\midrule
Qwen3-VL-4B-Thinking & Reasoning & 52.0 & N/R & N/R & 34/116 & 29.31 \\
Qwen3-VL-8B-Thinking & Reasoning & 50.0 & N/R & N/R & 42/116 & 36.21 \\
Qwen3-VL-32B-Thinking & Reasoning & 63.7 & N/R & N/R & 34/116 & 29.31 \\
GUI-Owl-1.5-8B-Thinking & Reasoning & 71.6 & 84/116 & 72.41 & 63/116 & 54.31 \\
GUI-Owl-1.5-32B-Thinking & Reasoning & N/P & N/R & N/R & 82/116 & 70.69 \\
\bottomrule
\end{tabular}}
\vspace{3pt}
\begin{minipage}{\textwidth}\scriptsize\raggedright
N/P: not published---the developer has not reported a score on the 116-task AndroidWorld set.
N/R: not run---no run was performed under that condition.
\end{minipage}
\end{table}

\clearpage
\section{AndroidFlux: Error Recovery Benchmark}
\label{error_recovery_ap}

\begin{table}[!t]
\centering
\footnotesize
\setlength{\tabcolsep}{3pt}
\caption{AndroidWorld clean-start success and human-verified recovery at the first-error
checkpoint. \emph{AndroidWorld} is the Docker run over the 116 AndroidWorld
tasks in Table~\ref{tab:appendix_aw}, given for reference. \emph{Scratch} is the same 116-task run
restricted to the 80 recovery tasks; $t_{\mathrm{all}}$ replays the
recorded prefix to the first-error checkpoint over the same 80 cases.}
\label{tab:aw-vs-recovery}
\resizebox{\textwidth}{!}{%
\begin{tabular}{llcccccc}
\toprule
& & \multicolumn{2}{c}{AndroidWorld (116)} & \multicolumn{2}{c}{Scratch (80)} & \multicolumn{2}{c}{$t_{\mathrm{all}}$ (80)} \\
\cmidrule(lr){3-4}\cmidrule(lr){5-6}\cmidrule(lr){7-8}
Model & Type & Correct & SR (\%) & Correct & SR (\%) & Correct & SR (\%) \\
\midrule
Qwen3-VL-2B-Instruct & General & 33/116 & 28.45 & 13/80 & 16.25 & 12/80 & 15.00 \\
Qwen3-VL-4B-Instruct & General & 51/116 & 43.97 & 21/80 & 26.25 & 21/80 & 26.25 \\
Qwen3-VL-8B-Instruct & General & 63/116 & 54.31 & 33/80 & 41.25 & 23/80 & 28.75 \\
Qwen3-VL-30B-A3B-Instruct & General & 27/116 & 23.28 & 11/80 & 13.75 & 21/80 & 26.25 \\
Qwen3-VL-32B-Instruct & General & 59/116 & 50.86 & 29/80 & 36.25 & 24/80 & 30.00 \\
Qwen3.6-27B (think on) & General & 32/116 & 27.59 & 14/80 & 17.50 & 25/80 & 31.25 \\
Qwen3.8-27B (think on) & General & 63/116 & 54.31 & 36/80 & 45.00 & 26/80 & 32.50 \\
Gemma-4-31B-IT (M3A) & General & 56/116 & 48.28 & 30/80 & 37.50 & 34/80 & 42.50 \\
Gemini Flash 3.0 (Qwen3-VL prompt) & General & 57/116 & 49.14 & 32/80 & 40.00 & 22/80 & 27.50 \\
Gemini Flash 3.5 (Qwen3-VL prompt) & General & 76/116 & 65.52 & 45/80 & 56.25 & 34/80 & 42.50 \\
Gemini Flash 3.8 & General & 72/116 & 62.07 & 41/80 & 51.25 & 43/80 & 53.75 \\
Gemini Pro 3.1 & General & 76/116 & 65.52 & 44/80 & 55.00 & 40/80 & 50.00 \\
GPT-5.6 Luna               & General & -- & -- & -- & -- & 27/80 & 33.75 \\
GPT-5.6 Terra   & General & -- & -- & -- & -- & 36/80 & 45.00 \\
\midrule
GUI-Owl-7B & Expert & 71/116 & 61.21 & 37/80 & 46.25 & 24/80 & 30.00 \\
GUI-Owl-32B & Expert & 75/116 & 64.66 & 43/80 & 53.75 & 21/80 & 26.25 \\
GUI-Owl-1.5-2B & Expert & 78/116 & 67.24 & 46/80 & 57.50 & 32/80 & 40.00 \\
GUI-Owl-1.5-4B & Expert & 74/116 & 63.79 & 40/80 & 50.00 & 27/80 & 33.75 \\
GUI-Owl-1.5-8B & Expert & 77/116 & 66.38 & 45/80 & 56.25 & 32/80 & 40.00 \\
GUI-Owl-1.5-32B & Expert & 77/116 & 66.38 & 45/80 & 56.25 & 26/80 & 32.50 \\
MAI-UI-2B & Expert & 44/116 & 37.93 & 17/80 & 21.25 & 17/80 & 21.25 \\
MAI-UI-8B & Expert & 60/116 & 51.72 & 30/80 & 37.50 & 28/80 & 35.00 \\
UI-TARS-1.0-7B-SFT & Expert & 31/116 & 26.72 & 13/80 & 16.25 & 13/80 & 16.25 \\
UI-TARS-1.5-7B & Expert & 25/116 & 21.55 & 10/80 & 12.50 & 19/80 & 23.75 \\
Gemini Flash 3.5 Computer Use & Expert & 45/116 & 38.79 & 31/80 & 38.75 & 28/80 & 35.00 \\
\midrule
Mobile-Agent-v3-7B & Agent & 78/116 & 67.24 & 47/80 & 58.75 & 31/80 & 38.75 \\
Mobile-Agent-v3-32B & Agent & 76/116 & 65.52 & 46/80 & 57.50 & 34/80 & 42.50 \\
MobileRun (Gemini Flash 3.0) & Agent & 95/116 & 81.90 & 63/80 & 78.75 & 54/80 & 67.50 \\
FinalRun (M3A + app knowledge) & Agent & 83/116 & 71.55 & 53/80 & 66.25 & 33/80 & 43.75 \\
\midrule
Qwen3-VL-4B-Thinking & Reasoning & 34/116 & 29.31 & 10/80 & 12.50 & 10/80 & 12.50 \\
Qwen3-VL-8B-Thinking & Reasoning & 42/116 & 36.21 & 18/80 & 22.50 & 13/80 & 16.25 \\
Qwen3-VL-32B-Thinking & Reasoning & 34/116 & 29.31 & 9/80 & 11.25 & 14/80 & 17.50 \\
GUI-Owl-1.5-8B-Thinking & Reasoning & 63/116 & 54.31 & 33/80 & 41.25 & 24/80 & 30.00 \\
GUI-Owl-1.5-32B-Thinking & Reasoning & 82/116 & 70.69 & 50/80 & 62.50 & 30/80 & 37.50 \\
\bottomrule
\end{tabular}
}\end{table}

\subsection{Human Verified Set}
Six source models each attempted 112 AndroidWorld tasks, yielding 672 trajectories in total. The 112 tasks were obtained by excluding four of the original 116 tasks due to environment issues that prevented reliable execution or evaluation. All 672 trajectories were then manually reviewed step by step.

We retained tasks for which at least one of the six source models produced a trajectory containing an error at a non-boundary step; trajectories whose only error occurred at the first or final step were excluded. From the eligible trajectories, we selected one source trajectory per task while balancing the source-model distribution as evenly as possible, resulting in 80 human-verified cases. Table~\ref{tab:hv-source} reports the resulting source-model distribution.

We then attempted to reconstruct each selected error state by replaying the recorded action sequence. For 65 cases, action replay successfully reproduced the annotated error state and was used directly for evaluation. For the remaining 15 cases, replaying the actions alone did not reproduce the intended error state, so we instead restore the corresponding state from an emulator snapshot. Details of the replay and snapshot procedure are provided in Appendix~\ref{app:replay-snapshot}. Table~\ref{tab:errorcat_human} reports the distribution of first-error categories and the corresponding category-wise recovery results.

\begin{table}[h]
\centering\scriptsize\setlength{\tabcolsep}{4pt}
\caption{Source-model distribution of the human-verified set. Replay cases are restored by
replaying the recorded action prefix; snapshot cases resume from a saved emulator snapshot.}
\label{tab:hv-source}
\begin{tabular}{lrrr}
\toprule
Source model & Replay ($n{=}65$) & Snapshot ($n{=}15$) & Total ($n{=}80$) \\
\midrule
Gemini Flash (Qwen3-VL prompt)  & 17 (26.15\%) & 5 (33.33\%) & 22 (27.50\%) \\
Qwen3-VL-2B         & 10 (15.38\%) & 4 (26.67\%) & 14 (17.50\%) \\
GUI-Owl-32B         & 10 (15.38\%) & 1 (6.67\%)  & 11 (13.75\%) \\
Mobile-Agent-v3-32B & 10 (15.38\%) & 1 (6.67\%)  & 11 (13.75\%) \\
MAI-UI-8B           &  9 (13.85\%) & 2 (13.33\%) & 11 (13.75\%) \\
Qwen3-VL-4B         &  9 (13.85\%) & 2 (13.33\%) & 11 (13.75\%) \\
\midrule
Total & 65 & 15 & 80 \\
\bottomrule
\end{tabular}
\end{table}

\subsection{$t{-}5$, $t{-}3$, $t{-}1$, $t{+}1^{+}$, and $t{+}1^{-}$}
\label{subset}

We construct two complementary extensions of the first-error checkpoint $t$.
The pre-error conditions $t{-}n$, for $n\in\{1,3,5\}$, measure how recovery
changes as the target policy is given more opportunity to act before the
annotated error.  For each condition, we replay the source trajectory through
step $t_{\mathrm{err}}-n$ and truncate the model-visible history at the same
point; the target policy then completes the task autonomously.  Thus, the
environment state and the primed history always correspond to the same source
prefix.  Some errors occur too early to admit a requested offset.  When
$t_{\mathrm{err}}<n$, we retain the task but map that condition to a clean
start, with neither replayed actions nor primed history.  This keeps all
conditions paired over the same 65 replay cases, but it also means that the
aggregate $t{-}3$ and $t{-}5$ scores mix true offset starts with clean starts.
We therefore report both the 65-case aggregate and, where relevant, the score
restricted to feasible cases.  Table~\ref{tab:t_n_construction} quantifies
this distinction: all cases admit $t{-}1$, whereas only 60.0\% and 32.3\%
admit $t{-}3$ and $t{-}5$, respectively.

\begin{table}[!h]
\centering
\small
\setlength{\tabcolsep}{5pt}
\caption{Construction of the pre-error conditions on the 65 replay-based,
human-verified cases.  A clipped case is evaluated from a clean start because
the source trajectory does not contain the requested pre-error offset.}
\label{tab:t_n_construction}
\begin{tabular}{lrrrr}
\toprule
Condition & Offset $n$ & Feasible & Clipped & Feasible share \\
\midrule
$t{-}1$ & 1 & 65 &  0 & 100.0\% \\
$t{-}3$ & 3 & 39 & 26 &  60.0\% \\
$t{-}5$ & 5 & 21 & 44 &  32.3\% \\
\bottomrule
\end{tabular}
\end{table}

\begin{table}[!h]
\centering
\small
\setlength{\tabcolsep}{5pt}
\caption{Recovery-action source distribution in the matched
$t{+}1^{+}$/$t{+}1^{-}$ subset.  Counts identify the models that authored the
selected actions, not the models that generated the original error trajectories.}
\label{tab:tplus1-action-sources}
\resizebox{\linewidth}{!}{%
\begin{tabular}{lrrr@{\hskip 14pt}lrrr}
\toprule
Recovery-action source & $+$ & $-$ & Total & Recovery-action source & $+$ & $-$ & Total \\
\midrule
MobileRun (Gemini Flash 3.0) & 9 & 1 & 10 & GUI-Owl-1.5-32B & 0 & 4 & 4 \\
Gemini Flash 3.5 Computer Use & 3 & 1 & 4 & GUI-Owl-1.5-8B (FinalRun app info) & 0 & 4 & 4 \\
GUI-Owl-1.5-2B & 3 & 1 & 4 & Gemini Flash 3.0 Computer Use & 0 & 4 & 4 \\
Gemma-4-31B-IT (M3A) & 3 & 1 & 4 & MAI-UI-2B & 1 & 2 & 3 \\
GUI-Owl-1.5-8B & 3 & 1 & 4 & MAI-UI-8B & 2 & 1 & 3 \\
GUI-Owl-1.5-8B-Thinking & 4 & 0 & 4 & Qwen3-VL-32B-Instruct & 1 & 2 & 3 \\
Gemini Flash 3.5 & 3 & 1 & 4 & Qwen3-VL-8B-Instruct & 2 & 1 & 3 \\
Mobile-Agent-v3-32B & 2 & 2 & 4 & Qwen3-VL-30B-A3B-Instruct & 1 & 2 & 3 \\
Gemini Flash 3.0 & 3 & 1 & 4 & Qwen3-VL-8B-Thinking & 1 & 2 & 3 \\
GUI-Owl-32B & 3 & 1 & 4 & UI-TARS-1.5-7B & 1 & 2 & 3 \\
GUI-Owl-7B & 1 & 3 & 4 & Qwen3-VL-2B-Instruct & 0 & 3 & 3 \\
GUI-Owl-1.5-4B & 2 & 2 & 4 & Qwen3-VL-4B-Instruct & 0 & 3 & 3 \\
Mobile-Agent-v3-7B & 1 & 3 & 4 & UI-TARS-1.0-7B-SFT & 0 & 3 & 3 \\
Qwen3-VL-4B-Thinking & 3 & 1 & 4 & & & & \\
\midrule
\textbf{Total (27 models)} & \textbf{52} & \textbf{52} & \textbf{104} &
\multicolumn{4}{c}{} \\
\bottomrule
\end{tabular}
}
\end{table}

The post-error conditions probe the downstream effect of one candidate recovery action.
For a given task, both members of a pair begin at the same restored error state
$t$.  We execute either the first action from a recovery rollout that
eventually succeeds, denoted $t{+}1^{+}$, or the first action from one that
eventually fails, denoted $t{+}1^{-}$; after this injected action, the evaluated
policy acts autonomously from the resulting state.  The superscripts therefore
describe the outcome of the rollout that supplied the action, rather than a
claim that the single action is sufficient for success or failure.

Candidate actions are collected from the 27 models with accepted $t$-condition
rollouts.  We retain a task only when both a successful-rollout action and a
failed-rollout action are available, producing 52 matched tasks (104 actions);
the remaining 13 tasks have no successful recovery in the candidate pool.
Before selection, actions are validated against the AndroidFlux action grammar,
and syntactically valid calls that cannot be executed in AndroidWorld are
discarded.  To limit dependence on any single policy, we process tasks with the
fewest candidates first and choose, on each side, the eligible contributor used
least often so far.  Table~\ref{tab:tplus1-action-sources} lists the agents
whose actions are selected for the resulting matched subset and their counts
on each side of the pair.

\subsection{LLM Verified Set}

Table~\ref{tab:lv-source} reports the source-model distribution. Coverage is deliberately unbalanced: the detector--verifier pipeline surfaces more usable first errors from weaker or more verbose policies, so a few sources dominate while the tail supplies breadth.

\begin{table}[!h]
\centering\scriptsize\setlength{\tabcolsep}{4pt}
\caption{Source-model distribution of the LLM-verified set ($n{=}312$ over 104 tasks,
21 source models).}
\label{tab:lv-source}
\begin{tabular}{lr@{\hskip 18pt}lr}
\toprule
Source model & Cases & Source model & Cases \\
\midrule
Gemini Flash (Qwen3-VL prompt)        & 72 (23.08\%) & GUI-Owl-1.5-2B          & 5 (1.60\%) \\
GUI-Owl-32B               & 49 (15.71\%) & Mobile-Agent-v3-7B      & 4 (1.28\%) \\
Gemma-4-31B-it-0716 (M3A) & 49 (15.71\%) & Qwen3-VL-4B             & 4 (1.28\%) \\
Gemma-4-31B (M3A)         & 42 (13.46\%) & Qwen3-VL-4B-thinking    & 3 (0.96\%) \\
Qwen3-VL-2B               & 24 (7.69\%)  & GUI-Owl-1.5-32B         & 3 (0.96\%) \\
GUI-Owl-7B                & 17 (5.45\%)  & GUI-Owl-1.5-4B          & 2 (0.64\%) \\
UI-TARS-1.0-7B-SFT        & 12 (3.85\%)  & MAI-UI-8B               & 2 (0.64\%) \\
GUI-Owl-1.5-8B (Think)    &  7 (2.24\%)  & Mobile-Agent-v3-32B     & 1 (0.32\%) \\
Qwen3-VL-32B-Instruct      &  7 (2.24\%)  & GUI-Owl-1.5-8B          & 1 (0.32\%) \\
MAI-UI-2B                 &  6 (1.92\%)  & Qwen3-VL-8B-thinking    & 1 (0.32\%) \\
                          &              & Qwen3-VL-30B-A3B-Instruct & 1 (0.32\%) \\
\midrule
\multicolumn{4}{r}{Total \quad 312 (100\%)} \\
\bottomrule
\end{tabular}
\end{table}

\begin{table}[!t]
\centering\scriptsize\setlength{\tabcolsep}{3pt}
\caption{LLM-verified set (312 cases), full table. \emph{Scratch} projects the model's 116-task
AndroidWorld clean-start run onto these 312 cases; $t$ replays the recorded prefix to the
first-error checkpoint. \emph{AndroidWorld} and \emph{s104} are the same 116-task run and its
projection onto the 104 distinct tasks this set covers, given for reference.}
\label{tab:appendix_lv}
\resizebox{\textwidth}{!}{%
\begin{tabular}{llrrrrrr}
\toprule
Model & Type & Scratch (312) & SR (\%) & $t$ (312) & SR (\%) & AndroidWorld (116) & s104 \\
\midrule
Qwen3-VL-2B-Instruct & General & 69/312 & 22.12 & 77/312 & 24.68 & 33/116 & 23/104 \\
Qwen3-VL-4B-Instruct & General & 120/312 & 38.46 & 101/312 & 32.37 & 51/116 & 40/104 \\
Qwen3-VL-8B-Instruct & General & 153/312 & 49.04 & 92/312 & 29.49 & 63/116 & 51/104 \\
Qwen3-VL-32B-Instruct & General & 141/312 & 45.19 & 106/312 & 33.97 & 59/116 & 47/104 \\
Gemma-4-31B-IT (M3A, think off) & General & 135/312 & 43.27 & 139/312 & 44.55 & 56/116 & 45/104 \\
Gemma-4-31B-IT (M3A, think on) & General & -- & -- & 148/312 & 47.44 & -- & -- \\
Gemini Flash 3.0 (Qwen3-VL prompt, think on) & General & 144/312 & 46.15  & 124/312 & 39.74 & -- & -- \\
Gemini 3.8 Flash & General & 183/312 & 58.65 & 192/312 & 61.54 & 72/116 & 61/104 \\
Gemini 3.1 Pro & General & 192/312 & 61.54 & 182/312 & 58.33 & 76/116 & 64/104 \\
Qwen3.6-27B (think on) & General & 72/312 & 23.08 & 105/312 & 33.65 & 32/116 & 24/104 \\
Qwen3.8-27B (think off) & General & 105/312 & 33.65 & 88/312 & 28.21 & 46/116 & 35/104 \\
Qwen3.8-27B (think on) & General & 156/312 & 50.00 & 118/312 & 37.82 & 62/116 & 52/104 \\
GPT-5.6 Luna               & General & -- & -- & 99/312  & 31.73 & -- & -- \\
GPT-5.6 Terra    & General & -- & -- & 133/312 & 42.63 & -- & -- \\

\midrule
GUI-Owl-7B & Expert & 177/312 & 56.73 & 85/312 & 27.24 & 71/116 & 59/104 \\
GUI-Owl-32B & Expert & 189/312 & 60.58 & 88/312 & 28.21 & 75/116 & 63/104 \\
GUI-Owl-1.5-8B & Expert & 195/312 & 62.50 & 138/312 & 44.23 & 77/116 & 65/104 \\
GUI-Owl-1.5-32B & Expert & 195/312 & 62.50 & 113/312 & 36.22 & 77/116 & 65/104 \\
MAI-UI-8B & Expert & 144/312 & 46.15 & 99/312 & 31.73 & 60/116 & 48/104 \\
UI-TARS-1.0-7B-SFT & Expert & 60/312 & 19.23 & 42/312 & 13.46 & 31/116 & 20/104 \\
Gemini Flash 3.5 Computer Use & Expert & 126/312 & 40.38 & 125/312 & 40.06 & 45/116 & 42/104 \\
\midrule
Mobile-Agent-v3-7B & Agent & 198/312 & 63.46 & 141/312 & 45.19 & 78/116 & 66/104 \\
Mobile-Agent-v3-32B & Agent & 192/312 & 61.54 & 144/312 & 46.15 & 76/116 & 64/104 \\
MobileRun (Gemini Flash 3.0) & Agent & 249/312 & 79.81 & 209/312 & 66.99 & 95/116 & 83/104 \\
MobileRun (Gemini Pro 3.1) & Agent & -- & -- & 215/312 & 68.91 & -- & -- \\
FinalRun (M3A + app knowledge) & Agent & 213/312 & 68.27 & 122/312 & 39.10 & 83/116 & 71/104 \\
\midrule
Qwen3-VL-32B-Thinking & Reasoning & 81/312 & 25.96 & 66/312 & 21.15 & 34/116 & 27/104 \\
GUI-Owl-1.5-32B-Thinking & Reasoning & 210/312 & 67.31 & 139/312 & 44.55 & 82/116 & 70/104 \\
GUI-Owl-1.5-8B-Thinking & Reasoning & 159/312 & 50.96 & 125/312 & 40.06 & 63/116 & 53/104 \\
Qwen3-VL-8B-Thinking   & Reasoning &  93/312 & 29.81 & 87/312 & 27.88  & 42/116 & 31/104 \\
Qwen3-VL-4B-Thinking   & Reasoning &  72/312 & 23.08 & 68/312 & 21.79  & 34/116 & 24/104 \\
\bottomrule
\end{tabular}}
\vspace{3pt}
\end{table}

\subsection{Human-Verified Set and LLM-Verified Set}

\begin{table}[!h]
\centering\scriptsize\setlength{\tabcolsep}{3pt}
\resizebox{\textwidth}{!}{%
\begin{tabular}{lcccccccccc ccccc}
\toprule
 & \multicolumn{10}{c}{Human-verified} & \multicolumn{5}{c}{LLM-verified} \\
\cmidrule(lr){2-11} \cmidrule(lr){12-16}
Model & Scratch & $t{-}5_{\text{rep}}$ & $t{-}3_{\text{rep}}$ & $t{-}1_{\text{rep}}$ & $t_{\text{rep}}$ & $t_{\text{snap}}$ & $t_{\text{all}}$ & $t{+}1^{+}$ & $t{+}1^{-}$ & $t{+}1_{\text{all}}$
      & Scratch & $t{-}1$ & $t$ & $t{+}1^{+}$ & $t{+}1^{-}$ \\
\midrule
Qwen3-VL-2B$^{\dagger}$ \citep{qwen3vl} & 16.25 & 13.85 & 16.92 & 23.08 & 12.31 & 26.67 & 15.00 & 28.85 & 17.31 & 23.08 & 22.12 & 27.56 & 24.68 & 41.83 & 32.69 \\
Qwen3-VL-4B$^{\dagger}$ \citep{qwen3vl} & 26.25 & 29.23 & 24.62 & 27.69 & 21.54 & 46.67 & 26.25 & 34.62 & 34.62 & 34.62 & 38.46 & 40.71 & 32.37 & 50.96 & 41.83 \\
GUI-Owl-32B \citep{guiowl}  & 53.75 & 40.00 & 43.08 & 35.38 & 21.54 & 46.67 & 26.25 & 28.85 & 34.62 & 31.73 & 60.58 & 41.35 & 28.21 & 41.35 & 42.79 \\
Mobile-Agent-v3-32B         & 57.50 & 44.62 & 49.23 & 47.69 & 41.54 & 46.67 & 42.50 & 44.23 & 48.08 & 46.15 & 61.54 & 51.28 & 46.15 & 67.31 & 55.29 \\
MAI-UI-8B \citep{maiui}     & 37.50 & 40.00 & 36.92 & 40.00 & 30.77 & 53.33 & 35.00 & 38.46 & 38.46 & 38.46 & 46.15 & 41.35 & 31.73 & 50.96 & 48.56 \\
Gemini Flash 3.0$^{\dagger}$           & 40.00 &  7.69 & 10.77 & 29.23 & 24.62 & 40.00 & 27.50 & 42.31 & 34.62 & 38.46 & 46.15 & 41.35 & 39.74 & 49.52 & 48.56 \\
\bottomrule
\end{tabular}
}
\caption{Human-verified (HV) and LLM-verified (LV) sets on error recovery over the HV set source agents.
Scratch runs from step 0 of all 80 HV tasks.
HV has 65 replay-based ($t_{\text{rep}}$) and 15 snapshot-based ($t_{\text{snap}}$) cases.
At $t{+}1$, $^{+}$ and $^{-}$ take the recovery action from the successful and failed recovery trajectory,
and $t{+}1_{\text{all}}$ aggregates both.
$^{\dagger}$Evaluated with the Qwen3-VL AndroidWorld prompt.}
\label{tab:hvcomparison}\label{tab:lvcomparison}
\end{table}

In Table~\ref{tab:hvcomparison}, the human-verified and LLM-verified sets show the same central pattern: performance degrades when agents resume from states containing an annotated first error. Before the error, a correct foreign prefix can still benefit some models: Qwen3-VL-2B and Qwen3-VL-4B outperform scratch from several $t{-}n$ checkpoints, suggesting that inherited progress can place the target in a useful state it may not reliably reach on its own while providing enough context to continue. Together, the $t{-}n$ checkpoints capture how inherited progress affects continuation before the first error, with $t{-}1$ providing the closest comparison to $t$.
The $t{+}1$ conditions show that a single recovery action can substantially alter downstream recoverability, depending on both the injected action and the recovering policy. 


\clearpage

\subsection{Cross-Agent Recovery Matrix}

\begin{table}[h]
\centering
\scriptsize
\setlength{\tabcolsep}{2.2pt}
\resizebox{\textwidth}{!}{%
\begin{tabular}{lrrrrrrr}
\toprule
Agent $\backslash$ Traj. Source
& Mobile-Agent-v3-32B
& Gemini Flash 3.0
& GUI-Owl-32B
& MAI-UI-8B
& Qwen3-VL-2B
& Qwen3-VL-4B
& Overall \\
\midrule
\multicolumn{8}{l}{\textbf{Source agents}} \\
Mobile-Agent-v3-32B & 30.0\% & 82.4\% & 30.0\% & 33.3\% & 10.0\% & 33.3\% & 41.5\% \\
Gemini Flash 3.0    & 10.0\% & 52.9\% & 10.0\% & 33.3\% & 0.0\%  & 22.2\% & 24.6\% \\
GUI-Owl-32B         & 20.0\% & 41.2\% & 10.0\% & 22.2\% & 0.0\%  & 22.2\% & 21.5\% \\
MAI-UI-8B           & 20.0\% & 52.9\% & 30.0\% & 33.3\% & 0.0\%  & 33.3\% & 30.8\% \\
Qwen3-VL-2B         & 0.0\%  & 41.2\% & 10.0\% & 0.0\%  & 0.0\%  & 0.0\%  & 12.3\% \\
Qwen3-VL-4B         & 10.0\% & 47.1\% & 20.0\% & 33.3\% & 0.0\%  & 0.0\%  & 21.5\% \\
\addlinespace
\multicolumn{8}{l}{\textbf{By class (Avg.)}} \\
General             & 28.8\% & 61.0\% & 23.8\% & 37.5\% & 10.0\% & 34.7\% & 35.6\% \\
Expert              & 20.0\% & 47.1\% & 26.7\% & 29.6\% & 13.3\% & 37.0\% & 30.8\% \\
Agentic framework   & 36.7\% & 68.6\% & 36.7\% & 51.9\% & 16.7\% & 55.6\% & 46.7\% \\
\bottomrule
\end{tabular}}
\caption{Error recovery evaluation matrix at condition $t$ on the human-verified set. Values are recovery success rates (\%). The upper block is the square source-to-source matrix. The lower block averages evaluated agents that are not trajectory sources, grouped by class as in Tables~\ref{tab:modelclass} and~\ref{tab:instreason}.}
\label{tab:matrix}
\end{table}

The matrix shows that recovery is governed strongly by which agent produced the error. Self-recovery is rarely the strongest entry in its own column: only Mobile-Agent-v3-32B recovers its own errors better than the other source agents, and Qwen3-VL-2B and Qwen3-VL-4B recover none of theirs, suggesting that the states on which an agent first fails are not necessarily the states it is best equipped to repair. Recovery also tracks the composition of first-error categories in Table~\ref{tab:errorcat_human}. The three sources with the largest Execution share, Gemini Flash 3.0, MAI-UI-8B, and Qwen3-VL-4B, are the three most recoverable under every agent class, with Gemini Flash 3.0 the most recoverable throughout. The remaining sources, Mobile-Agent-v3-32B, GUI-Owl-32B, and Qwen3-VL-2B, have at most $27.3\%$ Execution and at least $45.5\%$ Direction errors, and are the three hardest. An Execution error leaves the plan intact, so the target can often repair it by redoing the step, whereas a Direction error first requires recognizing that the route itself is wrong. Qwen3-VL-2B is the hardest source under every class, combining a Direction-heavy profile ($50.0\%$) with the highest Stagnation rate in the set ($28.6\%$). Agentic frameworks recover best from every source, for example reaching $55.6\%$ on Qwen3-VL-4B against $36.5\%$ and $33.3\%$ for general and expert agents. Overall, the matrix indicates that recoverability depends jointly on the source policy and the type of state induced by its first error. Further details on the error-category distributions are provided in Appendix~\ref{error_category}.

\clearpage

\subsection{Reproducibility}

Every configuration in Tables~\ref{tab:repro} and~\ref{tab:repro-flux} is run four times under
an identical setup: same Docker image, same emulator snapshot and pinned app versions, same
task list, and fixed seeds. This still does not make the pipeline deterministic. The emulator
renders asynchronously, so observations may be captured mid-transition before an animation,
network call, or background service has settled. Residual device state drifts with execution
order---caches, permission dialogs, and recency-ordered surfaces such as the launcher and file
pickers differ across nominally identical initial states---and clock-dependent verifiers are
sensitive to elapsed time. Batched serving further makes floating-point reduction order depend
on batch composition, so decoding is not bitwise reproducible even when greedy.

\begin{table}[h]
\centering\scriptsize\setlength{\tabcolsep}{4pt}
\caption{Repeated Docker runs on AndroidWorld ($116$ tasks). Cells report the number of
successful tasks with the corresponding success rate in parentheses.}
\label{tab:repro}
\resizebox{\textwidth}{!}{%
\begin{tabular}{lrrrrcc}
\toprule
& \multicolumn{4}{c}{Successful tasks (SR \%)} & \multicolumn{2}{c}{Aggregate} \\
\cmidrule(lr){2-5} \cmidrule(lr){6-7}
Model & Run 1 & Run 2 & Run 3 & Run 4 & Mean SR (\%) & Range (pp) \\
\midrule

Qwen3-VL-4B-Instruct  & 46 (39.66)& 46 (39.66)  & 46 (39.66)  & 47 (40.52)  & 39.88  & 0.86  \\
Qwen3-VL-8B-Instruct   & 66 (56.90) & 61 (52.59) & 60 (51.72) & 63 (54.31) & 53.88 & 5.17 \\
MAI-UI-8B              & 60 (51.72) & 60 (51.72) & 60 (51.72) & 60 (51.72) & 51.72 & 0.00 \\
GUI-Owl-7B             & 73 (62.93) & 71 (61.21) & 68 (58.62) & 71 (61.21) & 60.99 & 4.31 \\
GUI-Owl-1.5-4B  & 74 (63.79) & 74 (63.79)  & 74 (63.79)  & 74 (63.79)  & 63.79  & 0.00  \\
GUI-Owl-1.5-8B  & 77 (66.38) & 74 (63.79)  & 76 (65.52) & 79 (68.10) & 65.95  & 4.31  \\
\bottomrule
\end{tabular}}
\end{table}

\subsubsection{Replay and Snapshot}
\label{app:replay-snapshot}

The 80 human-verified cases are divided into 65 replay-based and 15 snapshot-based cases according to whether the annotated error state can be reproduced by action replay. For the 65 replay-based cases, we re-execute in the emulator the source model's recorded action sequence through the annotated first-error step $t$. A case is considered replayable when this procedure reaches the same erroneous state observed in the original trajectory. Once that state is reached, the recovering policy takes over and continues the task from the restored state. Its inherited history is constructed from the source trajectory metadata and rendered in the target policy's native format, as detailed in Appendix~\ref{app:generalized-history}.

For the remaining 15 cases, replaying the recorded actions from the task's initial state does not reproduce the original error state. A primary source of such mismatch is residual environment state from preceding tasks. For example, when the original AndroidWorld tasks are executed sequentially, previously used applications may change their positions in recency-ordered interfaces; replaying an individual task in isolation does not reproduce this preceding interaction history and can therefore lead to a different state despite executing the same actions. For these cases, we manually reconstruct the emulator to match the original error state and save it as a VM snapshot, allowing evaluation to resume directly from that state. The inherited history remains based on the original source trajectory, with only the minimal state-dependent details responsible for the replay mismatch adjusted to remain consistent with the reconstructed state. The recovering policy then receives this reconstructed state and history under the same evaluation protocol as in the replay-based cases.

\subsubsection{Recovery Trajectory Generation}
\label{app:recovery-gen}

Given a source trajectory $\tau^{\mathrm{src}}$ with annotated first error $t$, a recovery
trajectory is generated by instantiating the Resume operator of Eq.~\eqref{eq:resume} in the
live emulator at $k=t$, the checkpoint at which the error has been committed under an
otherwise correct prefix. The environment is first restored to $\hat{s}_t$ by re-executing the
recorded action prefix $a_{0:t}$ under the same task seed; for the $15$ human-verified
trajectories whose replay does not reach the error state
(Appendix~\ref{app:replay-snapshot}), the manually reconstructed emulator VM snapshot is
loaded instead. The target policy $\pi$ then receives the reconstructed context $h^{\pi}_t$,
built by re-rendering the schema-neutral step records of
Appendix~\ref{app:generalized-history} into $\pi$'s native prompt format, together with the
live observation of the restored screen. Nothing in that context marks the prefix as replayed,
erroneous, or authored by another model: from the target's perspective the episode is simply
in progress, and steps $0$ through $t$ read as its own past actions. The target is therefore
never told that a recovery is required---noticing the broken state is part of what AndroidFlux
measures.

From step $t+1$ the target acts autonomously over the action space of
Table~\ref{tab:action_space}, and actions are executed on the device exactly as in a
clean-start rollout. The episode terminates when the target emits a terminal action
(\texttt{status} or \texttt{answer}) or exhausts the remaining step budget $H-t$, and the
terminal state is scored by the task's programmatic verifier $r$. A trajectory is thus
credited only when the target both repairs the effect of $a_t$ and completes the remainder of
the task; partial repair without completion is scored as failure. When
$\pi=\pi_{\mathrm{src}}$ the run measures self-recovery, and otherwise the target inherits a
failure and a history authored by a different model---the cross-agent condition of
Table~\ref{tab:matrix}. Because the restored state and the rendered prefix are both fixed
before any target is run, the same broken state is handed identically to every agent in the
roster, so differences in success rate are attributable to the recovering policy alone.

Each (task, target) pair is run four times under the configuration of
Appendix~\ref{error_recovery_ap}. Table~\ref{tab:repro-flux} reports per-run recovery counts
for both the $65$ replay-restored cases and the additional $15$ snapshot cases. Run-to-run
spread on the replay-restored subset stays within $2$--$5$ recovered tasks per model,
comparable in absolute terms to the variation of clean-start AndroidWorld rollouts under the
same setup (Table~\ref{tab:repro}), so resuming from a restored state does not add instability
beyond what the environment itself contributes. Recovery rates broken down by first-error
category are reported in Appendix~\ref{error_category}.

\begin{table}[h]
\centering\scriptsize\setlength{\tabcolsep}{3pt}
\caption{Repeated Docker runs on the AndroidFlux human-verified set ($80$ tasks: $65$
action-replay, $15$ snapshot). Runs report the number of tasks recovered after resuming at
the first-error step $t$; mean columns report the mean count with the corresponding rate in
parentheses.}
\label{tab:repro-flux}
\resizebox{\textwidth}{!}{%
\begin{tabular}{lrrrrcrrrrccc}
\toprule
& \multicolumn{5}{c}{Action-replay ($65$)} & \multicolumn{5}{c}{Snapshot ($15$)} & \multicolumn{2}{c}{Overall ($80$)} \\
\cmidrule(lr){2-6} \cmidrule(lr){7-11} \cmidrule(lr){12-13}
Model & R1 & R2 & R3 & R4 & Mean (\%) & R1 & R2 & R3 & R4 & Mean (\%) & Mean (\%) & Range (pp) \\
\midrule
Qwen3-VL-4B-Instruct        & 12 & 14 & 15 & 13 & 13.50 (20.77) & 8 & 8 & 8 & 7 & 7.75 (51.67) & 21.25 (26.56) & 3.75 \\
Qwen3-VL-8B-Instruct        & 13 & 15 & 14 & 18 & 15.00 (23.08) & 6 & 7 & 3 & 6 & 5.50 (36.67) & 20.50 (25.63) & 8.75 \\
MAI-UI-8B  & 16 & 15 & 18 & 17 & 16.50 (25.38) & 7 & 7 & 8 & 7 & 7.25 (48.33) & 23.75 (29.69) & 5.00 \\
GUI-Owl-7B         & 20 & 14 & 18 & 17 & 17.25 (26.54) & 6 & 5 & 8 & 3 & 5.50 (36.67) & 22.75 (28.44) & 8.75 \\
GUI-Owl-1.5-4B     & 19 & 20 & 20 & 20 & 19.75 (30.38) & 8 & 6 & 5 & 7 & 6.50 (43.33) & 24.25 (30.31) & 2.50 \\
GUI-Owl-1.5-8B & 24 & 25 & 25 & 24 & 24.50 (37.69) & 8 & 8 & 9 & 8 & 8.25 (55.00) & 32.75 (40.94) & 2.5 \\
\bottomrule
\end{tabular}}
\end{table}

\paragraph{Episodes and models.}
The analysis uses the 312 episodes of the LLM-verified set. The 18 evaluated agents are listed in Table~\ref{tab:prefix_models}. GPT-5.6 Luna and Terra,
Gemma-4-31B-IT (M3A) and Mobile-Agent-v3-32B are not included.
A run counts as a success when its score is positive.
 
\paragraph{Gains and contrast.}
The unit is a (model, episode) pair, and the outcome of a run is 1 if it succeeds and 0 otherwise.
For each pair, the rewinding gain is the outcome under condition $t{-}1$ minus the outcome under
condition $t$, and the injection gain is the outcome under $t{+}1^{+}$ minus the outcome under $t{+}1^{-}$;
each is therefore $+1$, $0$ or $-1$. Averaged over the pairs in a bucket, a gain is the difference in
success rate between its two conditions, reported in percentage points.
Within a bucket, $\Delta$ is the injection gain minus the rewinding gain. The headline contrast is
$\Delta_{\ge 7} - \Delta_{\le 3}$: $\Delta$ in the latest bucket ($\ge 7$ steps before the error) minus
$\Delta$ in the earliest bucket ($\le 3$ steps), written ``latest $-$ earliest''.
 
\paragraph{Failure timing.}
The number of steps before the error is the index of the first erroneous step in the recorded source
trajectory, i.e.\ how many steps the source agent had executed before it went wrong.
Episodes are split into $\le 3$, $4$--$6$ and $\ge 7$ steps. These boundaries were fixed in an earlier
analysis of a subset of the data and were not tuned on the present results.
 
\paragraph{Coverage of the two gains.}
Rewinding is measured on all 312 episodes (116, 100 and 96 per bucket; the $n$ in Table~\ref{tab:prefix}),
i.e.\ 2{,}088, 1{,}800 and 1{,}728 model--episode pairs. The two $t{+}1$ arms each cover 208 episodes and
both arms are available for 196 of them, so injection is averaged over 196 episodes
(77, 64 and 55 per bucket; 1{,}386, 1{,}152 and 990 pairs).
 
\begin{table}[h]
\centering\footnotesize\setlength{\tabcolsep}{4pt}
\caption{Per-model $\Delta$ (injection $-$ rewinding, pp) by steps before the error, and the
latest $-$ earliest contrast. 17 of the 18 models are positive.}
\label{tab:prefix_models}
\begin{tabular}{llrrrr}
\toprule
Model & Type & $\le 3$ & $4$--$6$ & $\ge 7$ & Latest $-$ earliest \\
\midrule
Qwen3-VL-2B-Instruct            & General   &  $+0.02$ & $+14.62$ &  $+3.37$ &  $+3.35$ \\
Qwen3-VL-4B-Instruct            & General   & $-13.38$ &  $+6.75$ &  $+5.97$ & $+19.34$ \\
Qwen3-VL-8B-Instruct            & General   &  $-9.48$ &  $-6.88$ & $+10.64$ & $+20.13$ \\
Qwen3-VL-32B-Instruct           & General   &  $-7.34$ &  $-3.06$ & $-16.65$ &  $-9.30$ \\
Qwen3.8-27B (think on)          & General   & $-17.67$ &  $+5.81$ &  $+7.52$ & $+25.19$ \\
Gemini Flash 3.0 (think on)     & General   &  $-5.62$ &  $+5.81$ &  $-2.86$ &  $+2.76$ \\
Gemini Flash 3.8                & General   &  $+1.72$ &  $+5.12$ & $+11.70$ &  $+9.98$ \\
Gemini 3.1 Pro                  & General   &  $+0.01$ & $+13.19$ &  $+7.52$ &  $+7.51$ \\
\midrule
GUI-Owl-7B                      & Expert    & $-12.05$ & $-21.56$ &  $+8.05$ & $+20.10$ \\
GUI-Owl-32B                     & Expert    & $-17.25$ & $-24.81$ &  $-0.02$ & $+17.23$ \\
GUI-Owl-1.5-8B                  & Expert    &  $-9.92$ &  $+1.12$ & $+11.16$ & $+21.07$ \\
MAI-UI-8B                       & Expert    &  $-8.21$ & $-16.44$ &  $+1.80$ & $+10.01$ \\
UI-TARS-1.0-7B-SFT              & Expert    &  $-2.99$ &  $+2.94$ &  $-0.80$ &  $+2.19$ \\
Gemini Flash 3.5 Computer Use   & Expert    & $-16.82$ & $-14.62$ & $-13.54$ &  $+3.27$ \\
\midrule
Mobile-Agent-v3-7B              & Agent     & $-20.31$ & $+12.31$ & $+15.08$ & $+35.38$ \\
MobileRun (Gemini Flash 3.0)    & Agent     & $-12.95$ &  $-2.12$ &  $-8.31$ &  $+4.64$ \\
FinalRun (M3A + app knowledge)  & Agent     &  $-9.92$ &  $+3.69$ &  $+5.19$ & $+15.11$ \\
\midrule
GUI-Owl-1.5-8B-Thinking         & Reasoning & $-13.79$ & $+16.06$ &  $-1.82$ & $+11.97$ \\
\midrule
All 18 (pooled)                 &           &  $-9.77$ &  $-0.11$ &  $+2.44$ & $+12.22$ \\
\bottomrule
\end{tabular}
\end{table}
 
\paragraph{Uncertainty.}
Every model attempts every episode, so the pairs within an episode are correlated. We therefore resample
\emph{episodes}: the episodes of the earliest and the latest bucket are drawn with replacement, all pairs
of a drawn episode are kept, and the contrast is recomputed.
The 95\% interval is the percentile interval, $[+4.1, +20.8]$.
A one-way random-effects intra-class correlation with episodes as clusters is $0.090$ for rewinding and
$0.111$ for injection, with 18 pairs per episode. The design effects are $2.54$ and $2.89$, so the
effective sample is 39\% and 35\% of the nominal pairs.
We do not report a bootstrap $p$ for the headline contrast: it depends on only 6 to 13 of the
6{,}000 resamples, so it changes with the random seed, while the interval
changes by less than 0.25pp.
 
\paragraph{Robustness checks.}

Table~\ref{tab:prefix_robust} lists the checks cited in Section~6.4. Per model, the contrast is positive
for 17 of 18 models; the exception is Qwen3-VL-32B-Instruct ($-9.30$pp). A two-sided sign test gives
$p = 0.0001$. Difficulty quartiles split episodes by their mean success at $t$ over the 18 models
(Q1 hardest). For trajectory length, we cut at the 33rd and 67th percentiles of the recorded source trajectory
length; because many episodes have the same length, the three groups have 156, 54 and 102 episodes. Longer trajectories tend to have more steps before the error, so this check asks whether
trajectory length alone explains the contrast. For the random split, we shuffle the 312 episodes
once and split them into two halves of 156. Each group uses the same episode bootstrap with
4{,}000 resamples; $p$ is two-sided.
\begin{table}[h]
\centering\small\setlength{\tabcolsep}{5pt}
\caption{Latest $-$ earliest contrast (pp) within groups of episodes. All nine are positive.}
\label{tab:prefix_robust}
\begin{tabular}{llrrcr}
\toprule
Check & Group & Episodes & Contrast & 95\% CI & $p$ \\
\midrule
\multirow{4}{*}{Difficulty}
 & Q1 (hardest)  & 99 & $+12.69$ & $[+3.2, +22.8]$  & 0.010 \\
 & Q2            & 68 &  $+7.86$ & $[-9.5, +25.2]$  & 0.380 \\
 & Q3            & 69 & $+14.84$ & $[-4.1, +36.5]$  & 0.143 \\
 & Q4 (easiest)  & 76 & $+20.39$ & $[+4.1, +37.3]$  & 0.011 \\
\midrule
\multirow{3}{*}{Trajectory length}
 & $\le 10$ steps   & 156 & $+21.03$ & $[+2.5, +41.4]$  & 0.023 \\
 & $11$--$16$ steps &  54 & $+18.38$ & $[+0.9, +36.2]$  & 0.038 \\
 & $\ge 17$ steps   & 102 &  $+3.29$ & $[-9.9, +18.0]$  & 0.671 \\
\midrule
\multirow{2}{*}{Random split}
 & Half A & 156 & $+12.66$ & $[+1.2, +26.3]$ & 0.029 \\
 & Half B & 156 & $+11.49$ & $[+0.5, +23.3]$ & 0.044 \\
\bottomrule
\end{tabular}
\end{table}

\subsection{First Error and Recovery Across Error Categories}
\label{error_category}

\begin{figure}[h]
\centering

\begin{subfigure}[t]{0.32\textwidth}
    \centering
    \includegraphics[width=\linewidth]{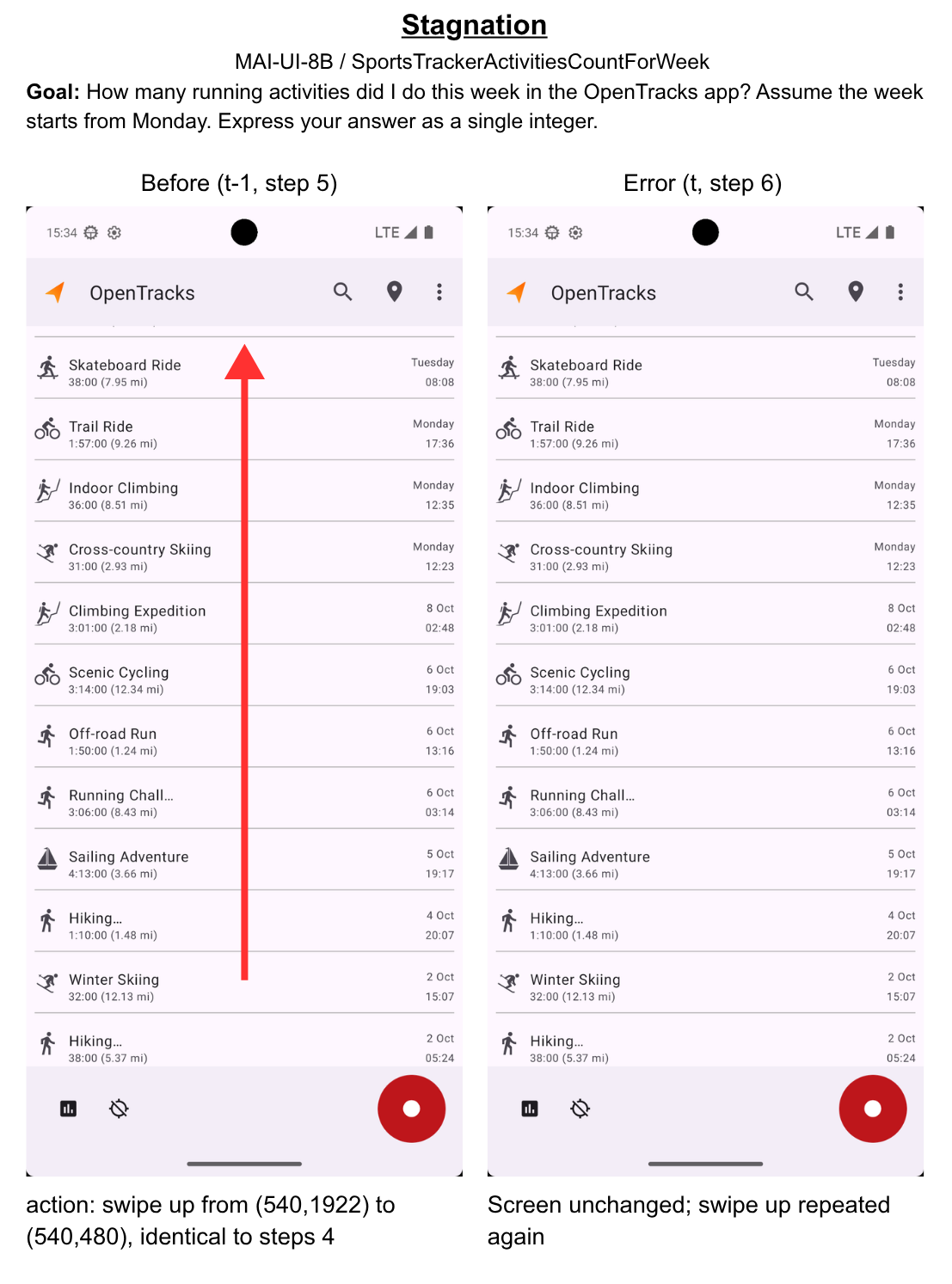}
    \caption{Stagnation}
    \label{fig:error_stagnation}
\end{subfigure}
\hfill
\begin{subfigure}[t]{0.32\textwidth}
    \centering
    \includegraphics[width=\linewidth]{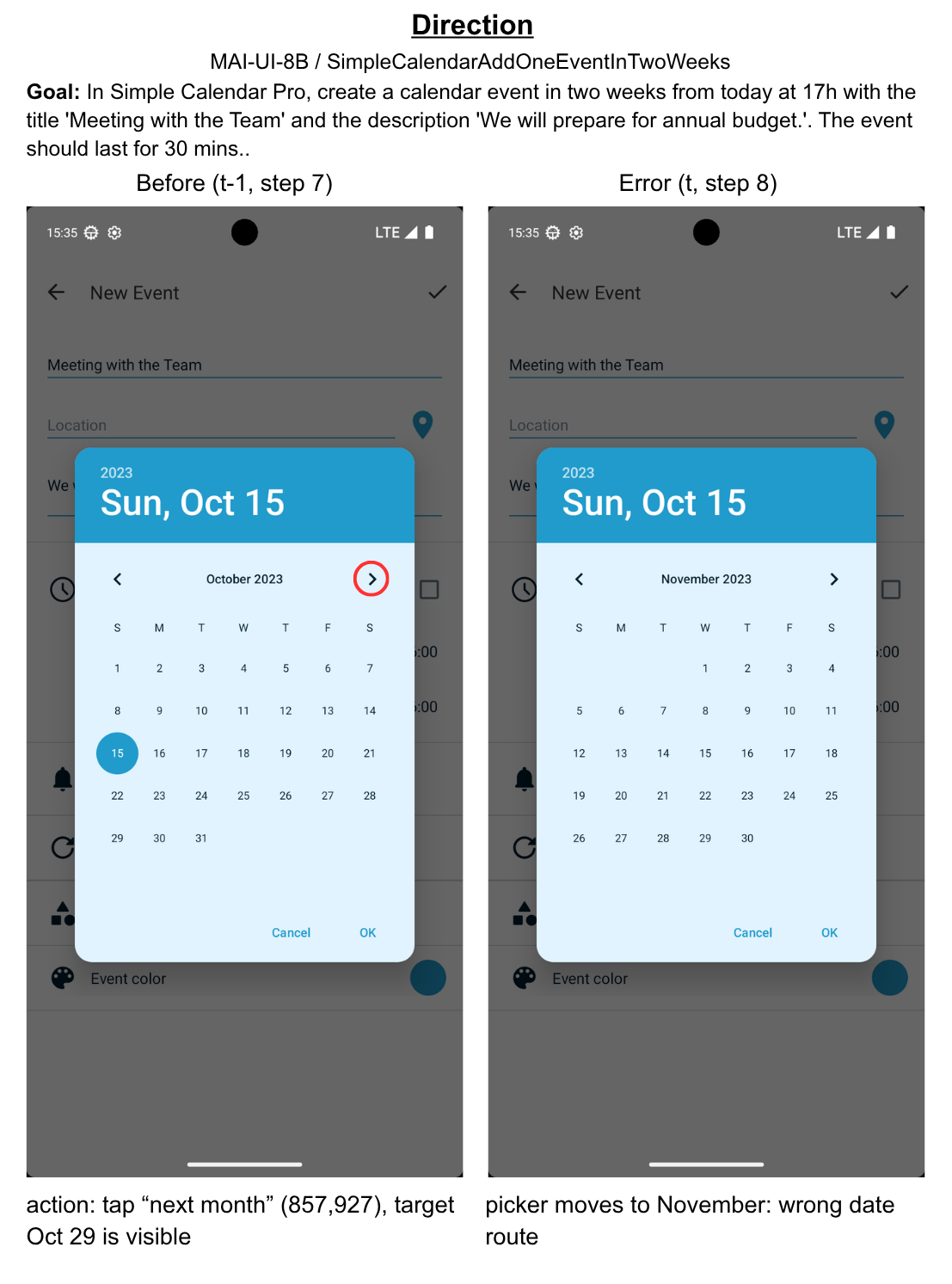}
    \caption{Direction}
    \label{fig:error_direction}
\end{subfigure}
\hfill
\begin{subfigure}[t]{0.32\textwidth}
    \centering
    \includegraphics[width=\linewidth]{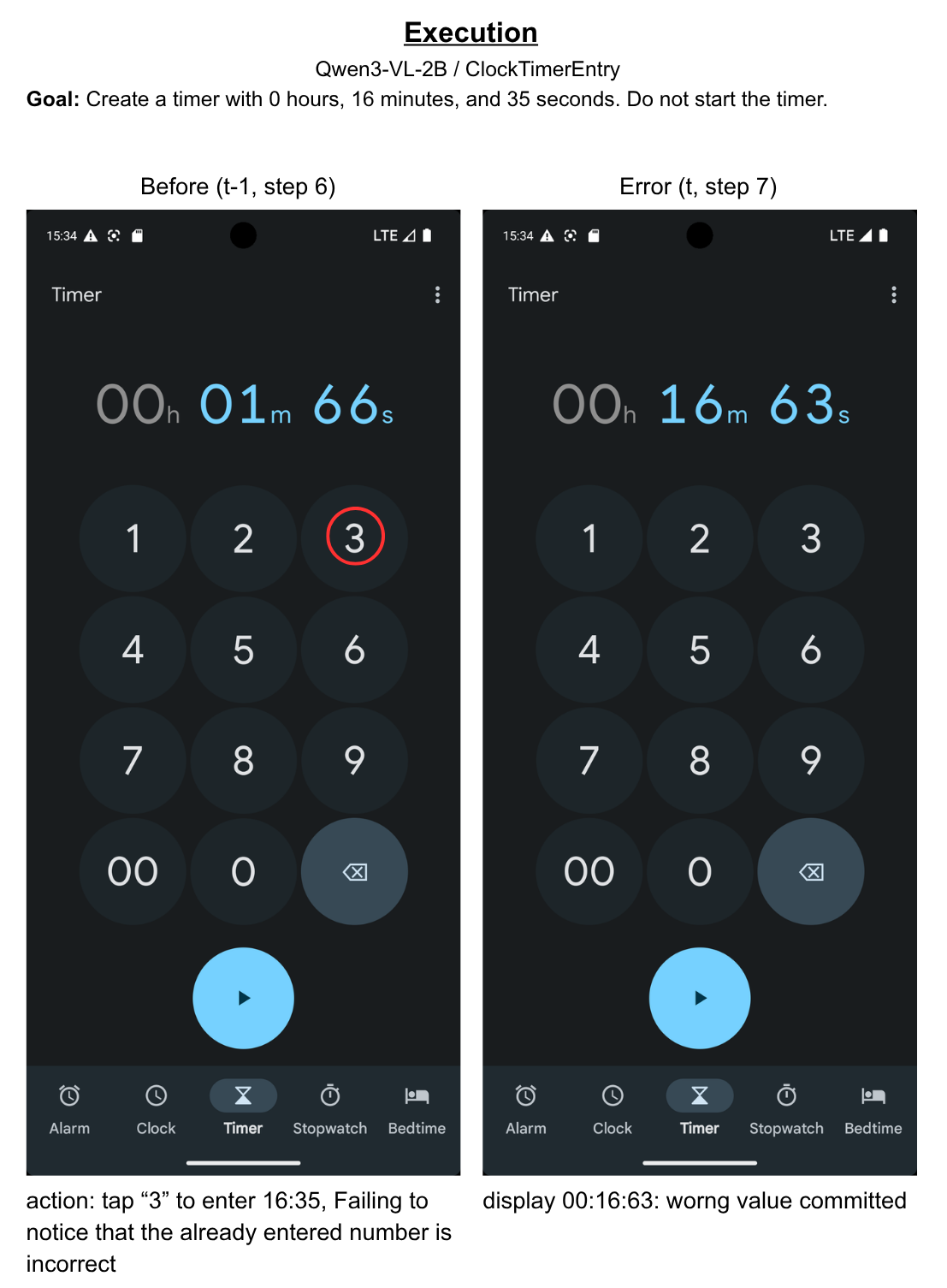}
    \caption{Execution}
    \label{fig:error_execution}
\end{subfigure}

\caption{One human-verified first-error example per class ($s_{t-1}$: screen before the action; $s_t$: screen where the error is observed). (a) Stagnation: an identical swipe is repeated on an unchanged list. (b) Direction: the action succeeds but leaves the goal route (October~29 was visible; the agent advances to November). (c) Execution: the correct control is targeted but the wrong value is committed (00:16:63 instead of 00:16:35).}
\label{fig:error_category}
\end{figure}

We categorize first errors by \emph{behavior} rather than by widget or app type.
Each step is viewed as a triple of (intended goal direction, chosen action, observed
screen transition), and the three classes are distinguished by which element of this
triple is broken. Classes are checked in a fixed priority order --- stagnation
$\rightarrow$ direction $\rightarrow$ execution --- so that a repeated action is
labeled as stagnation without further inspecting its direction or realization.
Figure~\ref{fig:error_category} shows, for each class, the screen
immediately before and after a representative first-error step
(a: stagnation, b: direction, c: execution), and
Tables~\ref{tab:errorcat_human} and~\ref{tab:errorcat_llm} summarize the
per-category counts for the two verified sets.

\paragraph{Stagnation.}
The agent re-issues an action that has already failed on an essentially unchanged
screen, or oscillates between states, so that the trajectory makes no visible
progress (Fig.~\ref{fig:error_stagnation}). The failure lies not in
\emph{what} the agent intends but in the fact that it repeats the attempt without
adapting. A first unsuccessful attempt is not counted: the error step is the point
at which repetition is established (typically the second or later identical
attempt), and actions with intrinsically subtle visual effects (e.g., text
editing, keyboard input) are excluded from the ``no screen change'' criterion.
Observed sub-patterns include identical re-taps on the same element, repeated
scrolling over an unchanged list, re-entering an already-failed route, and acting
on a nonexistent or already-achieved UI state.

\paragraph{Direction.}
The action executes successfully, but it moves the trajectory \emph{away} from the
goal (Fig.~\ref{fig:error_direction}): the agent commits to a visibly wrong
target, screen, date, or app, regresses by dismissing progress it has already
achieved, abandons a route on which the required affordance is already exposed, or
ignores a visible contradiction. Local execution is correct; the choice is not.
Exploratory route-opening, reversible entries into a wrong menu, and app switches
for source verification are protected as valid --- a direction error is assigned
only when a wrong target or state is committed, or when the agent leaves a
required screen without justification. This class subsumes goal-regressive,
goal-irrelevant, goal-direction, and unhandled system-error behaviors in our finer-grained
taxonomy.

\paragraph{Execution.}
The agent targets the right element with the right kind of action, but the
action's \emph{local effect} is wrong or absent
(Fig.~\ref{fig:error_execution}): the committed text, value, or selection
differs from the intent, the tap lands on a neighboring control, or a
launch/save/toggle produces no state change. The plan is correct; the realization
is not. The judgment unit is strictly the current step's intent versus its
immediate outcome --- an error is assigned when the immediately following
screenshot shows an incorrect committed value, but not merely because the overall
task remains unfinished, and stale UI or a missing toast after a correctly
targeted commit is not presumed to be a failure.

\paragraph{Boundary between stagnation and execution.}
``No screen change'' appears in both classes; the distinction is repetition.
The \emph{first} task-critical action that produces no effect (e.g., an app launch
or toggle that fails) is an execution error, whereas re-attempting an action that
has already failed is stagnation. Human-verified labels follow this convention
consistently.

\paragraph{Distributions across verified sets.}
Tables~\ref{tab:errorcat_human} and~\ref{tab:errorcat_llm} break down the
first-error class of every verified trajectory by the source model that produced
it: the first row of each table gives the overall class distribution of the set,
and the remaining rows show, for each source model, how many of its error
trajectories fall into each class. In the human-verified set, direction errors
are the largest class overall (43.75\%), and all three classes occur for every
source model; the balance shifts by model, with direction dominating for
GUI-Owl-32B and Qwen3-VL-2B, while MAI-UI-8B and Qwen3-VL-4B fail most often at
execution. In the LLM-verified set, stagnation is the largest class overall
(47.11\%) and execution the smallest (17.31\%), but this aggregate hides
substantial per-model variation: GUI-Owl-32B and the two Gemma-4-31B runs, which
together contribute 140 of the 312 trajectories, show a direction-dominant profile
(45--53\%) similar to the human-verified set, whereas the 72 Gemini Flash
trajectories are labeled as stagnation in 76.39\% of cases. Two caveats apply when
comparing the two sets: the LLM-verified labels are obtained by collapsing the
detector's five-way \texttt{pred\_error\_type} predictions into the three classes
above rather than by human judgment, and the model composition of the two sets
differs, so the per-model rows rather than the overall rows are the appropriate
unit of comparison.

\begin{table}[h]
\centering\scriptsize\setlength{\tabcolsep}{4pt}
\caption{First-error class distribution of the human-verified set ($n{=}80$), overall and by
source model. Percentages are within each row.}
\label{tab:errorcat_human}
\begin{tabular}{lrrrr}
\toprule
Source model & Cases & Stagnation & Direction & Execution \\
\midrule
All & 80 & 18 (22.50\%) & 35 (43.75\%) & 27 (33.75\%) \\
\midrule
Gemini Flash (Qwen3VL Prompt) & 22 & 4 (18.18\%) & 9 (40.91\%) & 9 (40.91\%) \\
Qwen3-VL-2B                   & 14 & 4 (28.57\%) & 7 (50.00\%) & 3 (21.43\%) \\
GUI-Owl-32B                   & 11 & 3 (27.27\%) & 6 (54.55\%) & 2 (18.18\%) \\
Mobile-Agent-v3-32B           & 11 & 3 (27.27\%) & 5 (45.45\%) & 3 (27.27\%) \\
MAI-UI-8B                     & 11 & 3 (27.27\%) & 3 (27.27\%) & 5 (45.45\%) \\
Qwen3-VL-4B                   & 11 & 1 (9.09\%)  & 5 (45.45\%) & 5 (45.45\%) \\
\bottomrule
\end{tabular}
\end{table}

\begin{table}[h]
\centering\scriptsize\setlength{\tabcolsep}{4pt}
\caption{First-error class distribution of the LLM-verified set ($n{=}312$, 21 source models),
overall and by source model. Classes are the detector's \texttt{pred\_error\_type} folded to the
three-way taxonomy (stagnation$\rightarrow$Stagnation; goal\_regression, goal\_irrelevant,
goal\_direction and system\_error$\rightarrow$Direction; incomplete\_execution$\rightarrow$Execution).
Percentages are within each row.}
\label{tab:errorcat_llm}
\begin{tabular}{lrrrr}
\toprule
Source model & Cases & Stagnation & Direction & Execution \\
\midrule
All & 312 & 147 (47.11\%) & 111 (35.58\%) & 54 (17.31\%) \\
\midrule
Gemini Flash (Qwen3VL Prompt) & 72 & 55 (76.39\%) &  8 (11.11\%) &  9 (12.50\%) \\
GUI-Owl-32B                   & 49 & 10 (20.41\%) & 25 (51.02\%) & 14 (28.57\%) \\
Gemma-4-31B-it-0716 (M3A)     & 49 & 12 (24.49\%) & 26 (53.06\%) & 11 (22.45\%) \\
Gemma-4-31B (M3A)             & 42 & 10 (23.81\%) & 19 (45.24\%) & 13 (30.95\%) \\
Qwen3-VL-2B                   & 24 & 18 (75.00\%) &  6 (25.00\%) &  0 (0.00\%)  \\
GUI-Owl-7B                    & 17 &  9 (52.94\%) &  5 (29.41\%) &  3 (17.65\%) \\
UI-TARS-1.0-7B-SFT            & 12 &  9 (75.00\%) &  3 (25.00\%) &  0 (0.00\%)  \\
GUI-Owl-1.5-8B (Think)        &  7 &  2 (28.57\%) &  3 (42.86\%) &  2 (28.57\%) \\
Qwen3-VL-32B-Instruct          &  7 &  6 (85.71\%) &  1 (14.29\%) &  0 (0.00\%)  \\
MAI-UI-2B                     &  6 &  1 (16.67\%) &  5 (83.33\%) &  0 (0.00\%)  \\
GUI-Owl-1.5-2B                &  5 &  1 (20.00\%) &  3 (60.00\%) &  1 (20.00\%) \\
Mobile-Agent-v3-7B            &  4 &  2 (50.00\%) &  2 (50.00\%) &  0 (0.00\%)  \\
Qwen3-VL-4B                   &  4 &  2 (50.00\%) &  2 (50.00\%) &  0 (0.00\%)  \\
Qwen3-VL-4B-thinking          &  3 &  3 (100.00\%) & 0 (0.00\%)  &  0 (0.00\%)  \\
GUI-Owl-1.5-32B               &  3 &  2 (66.67\%) &  1 (33.33\%) &  0 (0.00\%)  \\
GUI-Owl-1.5-4B                &  2 &  1 (50.00\%) &  1 (50.00\%) &  0 (0.00\%)  \\
MAI-UI-8B                     &  2 &  1 (50.00\%) &  1 (50.00\%) &  0 (0.00\%)  \\
Mobile-Agent-v3-32B           &  1 &  0 (0.00\%)  &  0 (0.00\%)  &  1 (100.00\%) \\
GUI-Owl-1.5-8B                &  1 &  1 (100.00\%) & 0 (0.00\%)  &  0 (0.00\%)  \\
Qwen3-VL-8B-thinking          &  1 &  1 (100.00\%) & 0 (0.00\%)  &  0 (0.00\%)  \\
Qwen3-VL-30B-A3B-Instruct     &  1 &  1 (100.00\%) & 0 (0.00\%)  &  0 (0.00\%)  \\
\bottomrule
\end{tabular}
\end{table}

\clearpage

\subsection{Generalized History and Context}
\label{app:generalized-history}

\begin{table}[!h]
\centering\scriptsize\setlength{\tabcolsep}{5pt}
\caption{The schema-neutral step record. Provenance: \emph{Copied} verbatim from the source step, \emph{Norm.}\ deterministically re-encoded, \emph{Synth.}\ generated when the source emits nothing usable, \emph{Annot.}\ from the detector--verifier labels. Only \emph{Synth.}\ fields reach a recovering agent; copied text is for auditing.}
\label{tab:canonical_fields}
\begin{tabular*}{\linewidth}{@{\extracolsep{\fill}}lllp{0.42\linewidth}@{}}
\toprule
Group & Field & Prov. & Content \\
\midrule
\texttt{step} & \texttt{memory\_step\_idx} & Norm. & Position in the replay prefix, contiguous from $0$ \\
& \texttt{source\_step\_idx} & Copied & Index in the original source trajectory \\
& \texttt{goal} & Copied & Task instruction $c$ in force at this step \\
\addlinespace
\texttt{screenshot} & \texttt{before\_screenshot} & Copied & Observation $o_i$ before the action \\
& \texttt{after\_screenshot} & Copied & Observation $o_{i+1}$ after the action \\
\addlinespace
\texttt{action} & \texttt{parsed\_action} & Norm. & Canonical tool call, coordinates rescaled to $[0,1]$ \\
& \texttt{json\_action\_type} & Norm. & Canonical type (\texttt{click}, \texttt{swipe}, \texttt{type}, \dots) \\
& \texttt{execution\_success} & Copied & Whether the emulator executed the action \\
& \texttt{action\_raw} & Copied & Source model's action string, unedited \\
\addlinespace
\texttt{thought} & \texttt{synthetic\_thought} & Synth. & Imperative restatement of local intent \\
& \texttt{thought\_raw} & Copied & Source model's reasoning $u_i$, unedited \\
\addlinespace
\texttt{summary} & \texttt{summary\_short} & Synth. & Past-tense one-liner; the history line a target reads \\
& \texttt{synthetic\_turn} & Synth. & Assistant turn in thinking/tool-call form \\
& \texttt{conclusion\_raw} & Copied & Source model's post-hoc summary, unedited \\
\addlinespace
\texttt{error} & \texttt{is\_error\_step} & Annot. & Step carries an error label \\
& \texttt{is\_selected\_error} & Annot. & Step is the first error $t$ \\
& \texttt{is\_before\_error} & Annot. & Step belongs to the correct prefix \\
\bottomrule
\end{tabular*}
\end{table}

\paragraph{Canonical record fields.}
Agents differ in what they emit alongside the action and in how they render history: window length, whether reasoning is retained, and summary format all vary with $\Phi_\pi$. Replaying a trajectory under a different policy therefore requires a canonical intermediate. Each recorded response is inverted to the schema-neutral record of Table~\ref{tab:canonical_fields}, then re-rendered in the target agent's native format, so no agent receives another agent's prompt format verbatim. Fields the source agent does not emit are synthesized, so every evaluated agent receives a complete history in the form it expects. The accessibility tree is deliberately not stored: agents that consume it receive the live tree from the restored state at resume time, since a recorded tree would describe the source trajectory's screen rather than the one the target is looking at.

Each episode wraps these steps with the task, source model, coordinate convention, replay stop $k$, and annotated first error $t$. Normalizing to a shared action schema decouples the record from the source model's coordinate convention, so each adapter re-emits the units its target expects. A record stores the full prefix and defers truncation to the adapter, which slices the latest steps under the target's own history length. One file holds a prebuilt variant per replay stop, so the $t{-}1$/$t$/$t{+}1$ conditions are materialized ahead of time rather than recomputed per run.

\paragraph{Synthesizing the absent fields.}
\texttt{summary} is what a recovering agent is most exposed to, so it is synthesized under a two-tier rule preferring authenticity but enforcing neutrality. We reuse the source model's own summary only if it is plain prose, opens with a past-tense verb, and survives sanitization: pronouns and role prefixes are stripped and purpose clauses truncated, since a stated purpose is the source model's belief about its own plan and would import that plan into another policy's context. Any candidate mentioning error, mistake, replay or recovery is discarded---a transcript narrating its own failure would tell the target that something has gone wrong, precisely the inference the benchmark measures. Otherwise we synthesize from an action-type template, of which \texttt{thought} is the imperative form. The filters are strict and most steps fall through, which is the conservative outcome: the history reads uniformly across source models, stating what was done without why or that it was wrong.

\paragraph{Re-rendering.}
Adapters follow the same principle, writing every replayed step so the inherited prefix reads as ordinary progress rather than as a recording. Records are prebuilt offline before any emulator time, alongside a projected model-ready view per target that makes each agent's context inspectable without running the benchmark. Because context is fixed before replay, a run falling back to an emulator snapshot receives an identical history, so the three conditions differ only in where the prefix stops.

\clearpage

\begin{figure*}[!h]
\centering
\begin{tcolorbox}[
    colback=white,
    colframe=black,
    boxsep=3pt,
    arc=0pt,
    outer arc=0pt,
    width=\textwidth
]
\textbf{(a) Generalized History and Context Generation Prompt}\\[3pt]

\begin{lstlisting}[
    basicstyle=\ttfamily\tiny,
    breaklines=true,
    breakatwhitespace=true,
    columns=fullflexible,
    keepspaces=true,
    showstringspaces=false
]
## BATCH IMAGE/TEXT ORDER
This batch contains source screenshots {FIRST_IDX} through {LAST_IDX}.
Only steps explicitly marked EVALUABLE below should appear in `step_evaluations`.
For each EVALUABLE step, compare its screenshot to the immediately following screenshot in this batch.
Exception: terminal/report actions, and final source steps whose following screenshot is not present, may be EVALUABLE without a following screenshot; judge them from the current screenshot, action marker, and prior batch context.
For `batch_summary`, describe each evaluated step with complete sentences for screen state, action taken, and visible result/claim. Use only screen/action/result fields and avoid terse label fragments.

## PRIOR BATCH SUMMARY HISTORY
These summaries describe earlier chronological source steps. Use them as history context only; the current batch images remain authoritative.

### Prior source steps {START}..{END}
{PRIOR_BATCH_SUMMARY}

### Prior source steps {START}..{END}
{PRIOR_BATCH_SUMMARY}

## CURRENT BATCH
The images and action text below are the current batch for source screenshots {FIRST_IDX}..{LAST_IDX}.

--- per-image text block, emitted after each screenshot ------------------------

[EVALUABLE step]
Image {POS}: source step {IDX} BEFORE screenshot with Step {IDX} action marker.
EVALUABLE source_step_idx: {IDX}
Immediate AFTER screenshot: Image {POS+1} / source step {IDX+1}
Markdown step index: {MD_STEP_IDX}
Action details:
{DISPLAY_ACTION}
Recent local intent: {ACTOR_THOUGHT}

[EVALUABLE terminal/report step with no following screenshot]
  the third line above is replaced by:
Terminal/report evaluation: no following screenshot exists for this final answer/terminate step. Evaluate whether the answer/completion claim is supported by this screenshot and the prior batch context; do not require a visual post-action transition.

[EVALUABLE final GUI step with no following screenshot]
  the third line above is replaced by:
Final-step current-only evaluation: no following screenshot exists for this final GUI action. Use the current screenshot, action marker, action text, and prior batch context to judge whether the action target/route/intent is visibly wrong or unsupported. Do not infer success or failure solely from the missing post-action screenshot.

[non-evaluable step]
Image {POS}: source step {IDX} {AFTER/context-only | context-only} screenshot.
Do not evaluate this image as a source_step_idx in this batch.

--- history field of the returned JSON object ----------------------------------

"batch_summary": "For every evaluated source step, write one or two complete natural-language sentences. Use this pattern, and make each field a full explanatory sentence rather than a short label: Step N: screen=<a full sentence describing the visible app/screen/state before the action>; action=<a full sentence describing what the agent attempted and why>; result=<a full sentence describing the visible change or completion/answer claim>. Include each evaluated source_step_idx explicitly."

Always include `batch_summary` even for short trajectories. The summary should be useful as history for the next batch, so it must mention both the visible screen state and the action taken for each evaluated step. Avoid terse fragments such as `screen=home screen`; write complete explanatory sentences inside each field.

When PRIOR BATCH SUMMARY HISTORY is provided, treat it as compact history from earlier source steps. Use it to understand previous progress, repeated failures, and carried-over state, but do not override visible evidence in the current batch.
\end{lstlisting}
\end{tcolorbox}

\caption{Prompt used to generate the generalized history and context. A batch of up to ten screenshots is interleaved with
one text block per step, and the model returns a \texttt{batch\_summary} describing every
evaluated step as a \texttt{screen}/\texttt{action}/\texttt{result} triple. That summary is
carried forward verbatim into the next batch's \texttt{PRIOR BATCH SUMMARY HISTORY} section, so
long trajectories are summarized incrementally rather than re-read in full.}
\label{fig:genhist_prompt}
\end{figure*}

\clearpage
\subsection{Detector--Verifier First Error Detection Framework}
\label{detection_framework}
Human annotation does not scale to a source-model pool of this size, so the LLM-verified set is labeled by a two-stage pipeline, both stages implemented with Gemini Flash 3.0. A detector reads a sliding window of ten step screenshots with the executed actions and proposes clustered error candidates; a verifier then re-checks each candidate against the resulting UI state and returns a single first-error step. We tuned both stages toward precision at the expense of recall, since a missed error only reduces coverage whereas a false label changes the state at which every downstream experiment begins.

Table~\ref{tab:detector-verifier} reports the outcome. Aggregated over six source models, the verifier raises step-level precision from 87.2 to 90.8 while leaving recall essentially unchanged, confirming that it removes false candidates rather than trading one error type for another. The per-source breakdown shows the operating point holding across the pool: step-level precision stays between 80.0 and 94.9 on every source model, whereas recall varies far more widely, from 37.8 on Mobile-Agent-v3-32B to 70.1 on Qwen3-VL-2B. Recall tracks how legible a model's failures are---weaker policies produce blunter, more detectable errors---but because coverage loss is the tolerated failure mode, this variation costs breadth rather than correctness. The one row that stands apart is Gemini Flash, whose high step-level precision (94.9) coincides with the lowest trajectory-level accuracy (55.0), indicating that its errors are correctly identified when found but more often missed at the trajectory level.

\begin{table}[!h]
\centering\scriptsize\setlength{\tabcolsep}{4pt}
\caption{First-error detection performance. The upper panel aggregates both pipeline stages over six source models ($N=666$ steps); the lower panel breaks the verifier stage down by source model.}
\label{tab:detector-verifier}
\begin{tabular*}{\linewidth}{@{\extracolsep{\fill}}lcccccc@{}}
\toprule
& \multicolumn{4}{c}{Step-level} & \multicolumn{2}{c}{Trajectory-level} \\
\cmidrule(lr){2-5}\cmidrule(lr){6-7}
& Precision & Recall & F1 & Accuracy & Precision & Accuracy \\
\midrule
\multicolumn{7}{l}{\emph{Pipeline stage (aggregate)}} \\
Detector & 87.2 & 51.6 & 64.9 & 94.9 & 89.8 & 70.7 \\
+ Verifier & 90.8 & 50.8 & 65.2 & 95.0 & 92.1 & 71.0 \\
\midrule
\multicolumn{7}{l}{\emph{By source model (verifier stage)}} \\
Mobile-Agent-v3-32B & 89.5 & 37.8 & 53.1 & 96.2 & 89.5 & 74.8 \\
Gemini Flash & 94.9 & 43.5 & 59.7 & 89.8 & 94.7 & 55.0 \\
GUI-Owl-32B & 90.6 & 56.9 & 69.9 & 96.7 & 90.6 & 78.4 \\
MAI-UI-8B & 80.0 & 41.7 & 54.8 & 96.0 & 90.9 & 73.0 \\
Qwen3-VL-2B & 94.7 & 70.1 & 80.6 & 94.8 & 94.7 & 77.5 \\
Qwen3-VL-4B & 88.6 & 48.4 & 62.6 & 94.3 & 88.6 & 67.6 \\
\bottomrule
\end{tabular*}
\end{table}

\clearpage
\subsection{Detector--Verifier Framework Prompt}
\begin{figure*}[!h]
\centering
\begin{tcolorbox}[
    colback=white,
    colframe=black,
    boxsep=3pt,
    arc=0pt,
    outer arc=0pt,
    width=\textwidth
]
\textbf{(a-1) Detector Prompt -- Part 1}\\[3pt]

\begin{lstlisting}[
    basicstyle=\ttfamily\tiny,
    breaklines=true,
    breakatwhitespace=true,
    columns=fullflexible,
    keepspaces=true,
    showstringspaces=false
]
You are the "Omniscient GUI Auditor". Evaluate a short sequence of Android GUI trajectory steps.

## GOAL
{GOAL}

## INPUT FORMAT
- You will receive up to 10 screenshots in chronological order.
- After each screenshot, you will receive a text block describing that screenshot and, when applicable, the action attempted from that screenshot.
- A step is evaluable only when both its BEFORE screenshot and the next screenshot are present in this batch.
- Exception: terminal/report steps such as answer or terminate may be evaluable without a following screenshot. Judge those claims from the current screenshot and prior batch context.
- Exception: a final GUI action may be evaluable without a following screenshot when the dataset ends there. Judge only whether the current screenshot, action marker, action text, and prior context make the action visibly wrong or unsupported; do not infer success/failure solely from the missing post-action screenshot.
- The last screenshot in the batch is AFTER/context only unless its text block explicitly says it is evaluable.
- Markers are drawn after converting model-output coordinates into actual screenshot pixel coordinates.
- For click/tap/long-press actions, the marker is where the agent attempted to touch.
- For swipe/drag/scroll actions, the arrow is the attempted gesture path.
- Screenshots are authoritative. If action text, intent text, or prior notes conflict with screenshot evidence, trust the screenshots.
- When PRIOR BATCH SUMMARY HISTORY is provided, treat it as compact history from earlier source steps. Use it to understand previous progress, repeated failures, and carried-over state, but do not override visible evidence in the current batch.

## MULTI-STEP EVALUATION POLICY
- Apply the conservative error-counting policy: judge visible GUI errors conservatively, but do not over-forgive task-critical failures.
- For each evaluable step, compare that step's screenshot with the immediately following screenshot.
- For terminal/report steps with no following screenshot, evaluate whether the answer or completion claim is supported by the current screenshot and prior batch context.
- For final GUI steps with no following screenshot, use current-only evidence: marker placement, visible target, stated intent, and prior context. Be conservative unless the wrong target/route/intent is visible.
- Judge the step's actual GUI transition, not merely whether the entire goal is unfinished.
- Do not mark a step as error just because later work remains.
- Do not use a later screenshot to punish an earlier step unless it is the immediate AFTER screenshot for that step.
\end{lstlisting}
\end{tcolorbox}

\caption{Prompt used by the first-stage detector to evaluate Android GUI trajectory steps (Part 1).}
\label{fig:detector_prompt_1}
\end{figure*}

\begin{figure*}[!h]
\centering
\begin{tcolorbox}[
    colback=white,
    colframe=black,
    boxsep=3pt,
    arc=0pt,
    outer arc=0pt,
    width=\textwidth
]
\textbf{(a-2) Detector Prompt -- Part 2 (continued)}\\[3pt]

\begin{lstlisting}[
    basicstyle=\ttfamily\tiny,
    breaklines=true,
    breakatwhitespace=true,
    columns=fullflexible,
    keepspaces=true,
    showstringspaces=false
]
- Reversible-action grace policy: a first reversible probe, route-opening action, harmless missed tap, or extra browsing action is valid if no wrong target/value/state is committed and it can be retried.
- A repeated failed route/no-op is an error only when the same target/route has already failed and the current transition again shows no material progress.
- A task-critical action can be an error on the first failed attempt when the required visible target is already present and the action should open/select/confirm/save/add/create/record/send/delete/type into it, but the immediate AFTER screenshot shows no useful change, a wrong screen/control, or a visible contradiction.
- For exact text/value/state requirements, mark an error when the immediate AFTER screenshot visibly shows the wrong committed text/value/state.
- After a correctly targeted save/delete/move/add/create/confirm action, do not infer failure solely from stale UI, delayed refresh, unchanged source lists, or missing toast messages. Require visible wrong state, wrong target, or explicit contradiction.
- If an inserted newline/blank line or text layout is visibly present in the screenshot, do not override that visual evidence with text history saying "No Change".
- For terminal/report-only actions, mark error only when visible UI or recent visual sequence directly proves the completion/answer is false, incomplete for visible required items, or contradicts an exact requested state.

## OUTPUT FORMAT
Return raw JSON only. Include exactly one object:
{
  "decision_rationale": "Brief evidence-based rationale. First summarize the visible screen/action progression in this batch, then state which steps are valid/errors and the visible evidence for each.",
  "batch_summary": "For every evaluated source step, write one or two complete natural-language sentences. Use this pattern, and make each field a full explanatory sentence rather than a short label: Step N: screen=<a full sentence describing the visible app/screen/state before the action>; action=<a full sentence describing what the agent attempted and why>; result=<a full sentence describing the visible change or completion/answer claim>. Include each evaluated source_step_idx explicitly.",
  "step_evaluations": [
    {
      "source_step_idx": 0,
      "is_error": false,
      "primary_verdict": "valid",
      "pred_error_type": null,
      "reason": "Immediate before/after evidence for this step."
    }
  ]
}

Rules for `primary_verdict`:
- `valid`: no visible current-step agent error.
- `error_stagnation`: repeated failed route, loop, or action based on nonexistent UI state with no progress.
- `error_direction`: visibly wrong target/route/state, regressive action, or ignored visible contradiction.
- `error_incomplete`: intended local effect visibly failed or produced an unusable partial result.

Rules for `pred_error_type`:
- Use `null` when `is_error=false`.
- Use one of `stagnation`, `goal_regression`, `goal_irrelevant`, `system_error`, or `incomplete_execution` when `is_error=true`.

Evaluate every source_step_idx that the text blocks mark as EVALUABLE. Do not invent extra step ids.
Always include `batch_summary` even for short trajectories. The summary should be useful as history for the next batch, so it must mention both the visible screen state and the action taken for each evaluated step. Avoid terse fragments such as `screen=home screen`; write complete explanatory sentences inside each field.
\end{lstlisting}
\end{tcolorbox}

\caption{Prompt used by the first-stage detector to evaluate Android GUI trajectory steps (Part 2).}
\label{fig:detector_prompt_2}
\end{figure*}

\begin{figure*}[p]
\centering
\begin{tcolorbox}[
    colback=white,
    colframe=black,
    boxsep=3pt,
    arc=0pt,
    outer arc=0pt,
    width=\textwidth
]

\textbf{(b-1) Cluster Verifier Prompt -- Part 1}\\[3pt]

\begin{lstlisting}[
    basicstyle=\ttfamily\tiny,
    breaklines=true,
    breakatwhitespace=true,
    columns=fullflexible,
    keepspaces=true,
    showstringspaces=false
]
You are the "Omniscient GUI Auditor". Evaluate a short sequence of Android GUI trajectory steps.

## GOAL
{GOAL}

## INPUT FORMAT
- You will receive up to 10 screenshots in chronological order.
- After each screenshot, you will receive a text block describing that screenshot and, when applicable, the action attempted from that screenshot.
- A step is evaluable only when both its BEFORE screenshot and the next screenshot are present in this batch.
- Exception: terminal/report steps such as answer or terminate may be evaluable without a following screenshot. Judge those claims from the current screenshot and prior batch context.
- Exception: a final GUI action may be evaluable without a following screenshot when the dataset ends there. Judge only whether the current screenshot, action marker, action text, and prior context make the action visibly wrong or unsupported; do not infer success/failure solely from the missing post-action screenshot.
- The last screenshot in the batch is AFTER/context only unless its text block explicitly says it is evaluable.
- Markers are drawn after converting model-output coordinates into actual screenshot pixel coordinates.
- For click/tap/long-press actions, the marker is where the agent attempted to touch.
- For swipe/drag/scroll actions, the arrow is the attempted gesture path.
- Screenshots are authoritative. If action text, intent text, or prior notes conflict with screenshot evidence, trust the screenshots.
- When PRIOR BATCH SUMMARY HISTORY is provided, treat it as compact history from earlier source steps. Use it to understand previous progress, repeated failures, and carried-over state, but do not override visible evidence in the current batch.

## MULTI-STEP EVALUATION POLICY
- Apply the conservative error-counting policy: judge visible GUI errors conservatively, but do not over-forgive task-critical failures.
- For each evaluable step, compare that step's screenshot with the immediately following screenshot.
- For terminal/report steps with no following screenshot, evaluate whether the answer or completion claim is supported by the current screenshot and prior batch context.
- For final GUI steps with no following screenshot, use current-only evidence: marker placement, visible target, stated intent, and prior context. Be conservative unless the wrong target/route/intent is visible.
- Judge the step's actual GUI transition, not merely whether the entire goal is unfinished.
- Do not mark a step as error just because later work remains.
- Do not use a later screenshot to punish an earlier step unless it is the immediate AFTER screenshot for that step.
- Reversible-action grace policy: a first reversible probe, route-opening action, harmless missed tap, or extra browsing action is valid if no wrong target/value/state is committed and it can be retried.
- A repeated failed route/no-op is an error only when the same target/route has already failed and the current transition again shows no material progress.
- A task-critical action can be an error on the first failed attempt when the required visible target is already present and the action should open/select/confirm/save/add/create/record/send/delete/type into it, but the immediate AFTER screenshot shows no useful change, a wrong screen/control, or a visible contradiction.
- For exact text/value/state requirements, mark an error when the immediate AFTER screenshot visibly shows the wrong committed text/value/state.
- After a correctly targeted save/delete/move/add/create/confirm action, do not infer failure solely from stale UI, delayed refresh, unchanged source lists, or missing toast messages. Require visible wrong state, wrong target, or explicit contradiction.
- If an inserted newline/blank line or text layout is visibly present in the screenshot, do not override that visual evidence with text history saying "No Change".
- For terminal/report-only actions, mark error only when visible UI or recent visual sequence directly proves the completion/answer is false, incomplete for visible required items, or contradicts an exact requested state.

## CLUSTER VERIFIER: FIRST-COUNTED-ERROR REFINEMENT
The first-stage multi-step detector marked this consecutive source-step run as errors:
[{CLUSTER}]
Cluster length: {N}
All first-stage predicted-error steps visible in this local verifier window:
{PRED_ERROR_STEPS_IN_WINDOW}

You will receive screenshots and action text around that run.
Your job is NOT root-cause analysis. Your job is to refine the detector output
under the same conservative error-counting policy used by the first-stage detector.

Select at most one candidate step from [{CLUSTER}]:
\end{lstlisting}

\end{tcolorbox}

\caption{Cluster verifier prompt for first-counted-error refinement (Part 1).}
\label{fig:cluster_verifier_prompt_1}
\end{figure*}

\begin{figure*}[p]
\centering
\begin{adjustbox}{minipage=\linewidth,max totalsize={\linewidth}{0.92\textheight}}
\begin{tcolorbox}[
    colback=white,
    colframe=black,
    boxsep=3pt,
    arc=0pt,
    outer arc=0pt,
    width=\textwidth
]

\textbf{(b-2) Cluster Verifier Prompt -- Part 2 (continued)}\\[3pt]

\begin{lstlisting}[
    basicstyle=\ttfamily\tiny,
    breaklines=true,
    breakatwhitespace=true,
    columns=fullflexible,
    keepspaces=true,
    showstringspaces=false
]
- Choose the first candidate that qualifies as the first counted error under the conservative error-counting policy.
- Do NOT choose the earliest causal-looking step merely because it starts an inefficient route.
- Do NOT choose a candidate merely because the full task is unfinished.
- Do NOT choose a first reversible probe, first no-op tap/scroll, first wrong-but-recoverable menu route,
  first source check, first app switch, or first plausible retry. These are protected by the reversible-action grace policy.
- For stagnation/repetition clusters, keep only the first candidate where repetition is already established
  and the current step itself is no longer a protected probe/retry. If an earlier candidate is just the first
  failed attempt and a later candidate is the true repeated failure, choose the later candidate.
- Be stricter for repetition than the first-stage detector: a second similar action is often still a protected
  retry. Prefer the later repeated failure when the cluster shows a sequence such as first attempt -> second
  attempt -> third/no-progress confirmation. Do not select the first or second item of a repetition cluster
  unless the current screenshot/action visibly commits an irreversible wrong state.
- For long app-switch/source-check loops, do not choose an early app switch or early source reopening just
  because the loop is inefficient. Prefer the point where the loop is clearly established after at least two
  full repeated cycles, or return null if the counted error appears later than the candidate evidence.
- For information-search tasks, do not treat back-and-forth scrolling/checking as an error unless the local
  evidence clearly shows a repeated no-progress loop or a visibly false final answer.
- For answer/terminate candidates in a clean or information-search trajectory, return null unless the current
  screenshot or immediate prior visual context directly proves the answer is false. Do not use broad memory,
  hidden/off-screen items, or uncertain interpretation of earlier screens to turn a clean trajectory into an error.
- For clean/recoverable paths, if later context shows the agent continues useful progress or recovers without
  committing a wrong state, return null.
- If a candidate looks wrong but a later context image shows the agent returns to a useful state or continues
  the intended workflow, treat the candidate as recoverable and return null.
- If the true counted error appears to be outside the candidate cluster, return null. Do not force a choice.
- If none of the candidate steps qualifies as a counted error under the conservative error-counting policy, return null.
- If a later candidate is the first step that clearly satisfies the error criterion after several protected
  setup/probe/retry steps, prefer the later candidate over earlier plausible causes. This verifier corrects
  over-eager early error labels.
- Do not let the grace policy erase every candidate when the later candidate is the first repeated no-progress
  step. In a short two-step cluster where the first tap/click is a protected no-op probe and the second repeats
  the same intended action with no visible progress, choose the second step as the first counted stagnation error.
- For tasks that require an exact target value or state, an approximate or exploratory selection is not protected
  once the agent commits to a visibly inconsistent value or state, or abandons visible evidence of the required
  value/state in favor of an unsupported assumption. If an earlier candidate is exploratory and a later candidate
  makes that commitment, choose the later candidate.
- If the local window includes a nearby non-cluster predicted-error step, you may list it in
  `drop_additional_pred_error_steps` only when the evidence clearly shows it is protected/valid. This is only for
  removing obvious local false positives; do not add new error steps outside the cluster.

Use screenshots as authoritative evidence. Action text, prior detector predictions, and detector reasons are
only hints. Prefer null over an early over-aggressive choice when the evidence is ambiguous.

## OUTPUT FORMAT (JSON ONLY)
{
  "decision_rationale": "Brief evidence-based rationale: compare the candidate steps under the conservative error-counting policy and explain why earlier candidates are protected or why one candidate is the first counted error.",
  "predicted_error_cluster": [{CLUSTER}],
  "core_error_step": integer | null,
  "keep_error_steps": [integer],
  "drop_error_steps": [integer],
  "drop_additional_pred_error_steps": [integer],
  "confidence": "high" | "medium" | "low",
  "policy_basis": "first_counted_error" | "all_candidates_protected_or_recoverable" | "true_error_outside_cluster" | "insufficient_evidence",
  "reason": "Why the selected step is the first counted error, or why no candidate should be kept."
}
\end{lstlisting}

\end{tcolorbox}
\end{adjustbox}

\caption{Cluster verifier prompt for first-counted-error refinement (Part 2).}
\label{fig:cluster_verifier_prompt_2}
\end{figure*}

\clearpage
\section{AndroidFlux: Recovery Action Reward Modeling Benchmark}
\label{rm_apd}

The recovery-action reward modeling benchmark asks whether a reward model can rank two
actions proposed at the same AndroidWorld checkpoint by their downstream recoverability,
with the preference grounded in environment outcomes rather than local heuristics or an
LLM judge. Construction proceeds in five stages: source trajectories are generated and
their first errors annotated, checkpoints are restored from them, candidate recovery
actions are collected into a group at each checkpoint, a fixed panel rolls out from every
executed candidate, and the resulting outcome margin determines the released pair.

\paragraph{Source trajectories.}
Checkpoints are restored from trajectories authored by ten agents spanning general,
expert, and agentic-framework classes.
Table~\ref{tab:rm_source_composition}\subref{tab:rm_source_trajectory_distribution}
gives the distribution over the 100 released cases; this provenance is separate from the
recovery-action generators of Table~\ref{tab:rm_roster}, which author the candidates
compared at a checkpoint rather than the trajectory that produced it.

\paragraph{Checkpoint and error type.}
Each checkpoint is restored either at a first error annotated by the detector--verifier
pipeline of Appendix~\ref{detection_framework} or at a clean intermediate state.
Table~\ref{tab:rm_source_composition}\subref{tab:rm_error_distribution} reports the
breakdown: 72 of the 100 checkpoints carry a detected error and 28 are clean-path
controls.
The detector's fine-grained labels are categorized into the taxonomy of
Appendix~\ref{error_category}: \emph{stagnation} remains Stagnation;
\emph{goal regression} and \emph{goal irrelevant} form Direction; and
\emph{incomplete execution} forms Execution. Direction is the most frequent detected
error at 43.06\%, and all three categories are represented in double digits, so no single
failure mode dominates the error-path half of the set.

\begin{table}[h]
\caption{Provenance of the 100 restored checkpoints. Shares in (b) are normalized over all
100 cases and over the 72 detected-error cases respectively.}
\label{tab:rm_source_composition}
\centering\scriptsize
\begin{subtable}[b]{0.54\linewidth}
\caption{Source-trajectory agent.}
\label{tab:rm_source_trajectory_distribution}
\centering\setlength{\tabcolsep}{4pt}
\begin{tabular}{lrr}
\toprule
Source-trajectory agent & Cases & Share \\
\midrule
Qwen3-VL-8B-Instruct   & 24 & 24.00\% \\
Qwen3-VL-4B-Instruct   & 24 & 24.00\% \\
MAI-UI-8B              & 17 & 17.00\% \\
Qwen3-VL-2B-Instruct   & 13 & 13.00\% \\
Qwen3-VL-32B-Instruct  &  7 &  7.00\% \\
Mobile-Agent-v3-7B     &  7 &  7.00\% \\
GUI-Owl-1.5-2B         &  4 &  4.00\% \\
GUI-Owl-1.5-8B         &  2 &  2.00\% \\
GUI-Owl-1.5-32B        &  1 &  1.00\% \\
GUI-Owl-7B             &  1 &  1.00\% \\
\midrule
Total & 100 & 100.00\% \\
\bottomrule
\end{tabular}

\end{subtable}%
\hfill
\begin{subtable}[b]{0.44\linewidth}
\caption{Checkpoint and error type.}
\label{tab:rm_error_distribution}
\centering\setlength{\tabcolsep}{4pt}
\begin{tabular}{lrrr}
\toprule
Checkpoint type & Cases & \% 100 & \% err. \\
\midrule
Clean path         & 28 & 28.00\% & -- \\
\addlinespace
Direction          & 31 & 31.00\% & 43.06\% \\
Execution          & 23 & 23.00\% & 31.94\% \\
Stagnation         & 18 & 18.00\% & 25.00\% \\
\midrule
Detected error     & 72 & 72.00\% & 100.00\% \\
Total              & 100 & 100.00\% & -- \\
\bottomrule
\end{tabular}

\end{subtable}

\end{table}

\paragraph{Candidate group generation.}
At each restored checkpoint $g$, the five generators marked in
Table~\ref{tab:rm_roster} independently propose their first action, forming the
candidate group compared at that state. Every candidate is canonicalized into the shared
AndroidFlux action grammar of Table~\ref{tab:action_space}, so model-specific tool syntax
and coordinate conventions cannot reveal the generator, and invalid or non-executable
calls are discarded.

\paragraph{Recovery rollouts from $t{+}1$.}
Each surviving candidate $a$ is executed from $\hat{s}_g$, placing the environment at the
$t{+}1$ state that action induces. The fixed twelve-agent panel $\mathcal{P}$ of
Table~\ref{tab:rm_roster} then resumes independently from that state under the protocol
of Appendix~\ref{app:recovery-gen}, and the candidate's outcome score $S_g(a)$ is the
number of panel agents that reach the task goal. Because every candidate in a group is
scored by the same panel from the same checkpoint, the scores differ only in the action
that produced the $t{+}1$ state.

\begin{table}[h]
\centering\scriptsize\setlength{\tabcolsep}{10pt}
\caption{Model roster of the reward modeling benchmark. \emph{Cand.} marks the five
policies that propose candidate recovery actions at a checkpoint; \emph{Panel} marks the
twelve continuation agents that roll out from each executed candidate. MAI-UI panel
entries are history-conditioned and Gemini Flash 3.5 Computer Use is run task-only.}
\label{tab:rm_roster}
\begin{tabular}{lcc@{\hskip 26pt}lcc}
\toprule
Model & Cand. & Panel & Model & Cand. & Panel \\
\midrule
Qwen3-VL-2B-Instruct & \cmark & \cmark & MAI-UI-2B & -- & \cmark \\
Qwen3-VL-4B-Instruct & --     & \cmark & MAI-UI-8B & \cmark & \cmark \\
Qwen3-VL-8B-Instruct & \cmark & \cmark & Gemini Flash 3.0 (M3A) & -- & \cmark \\
GUI-Owl-7B           & \cmark & \cmark & MobileRun (Gemini Flash 3.0) & -- & \cmark \\
GUI-Owl-1.5-4B       & --     & \cmark & Gemini Flash 3.5 Computer Use & -- & \cmark \\
GUI-Owl-1.5-8B       & \cmark & \cmark & & & \\
GUI-Owl-1.5-8B-Thinking & --  & \cmark & \textbf{Total} & \textbf{5} & \textbf{12} \\
\bottomrule
\end{tabular}
\end{table}

\paragraph{Pair selection and outcome margin.}
Selection runs independently within each group, taking the highest-scoring candidate as
the positive action $a^{+}_g$ and the lowest-scoring as the negative $a^{-}_g$. The two
canonicalized actions must differ, so the same action proposed by two models cannot
receive opposite labels. When several candidates occupy one success tier we favor the
generator used least often in that role so far, which improves source coverage without
overriding the panel outcome; both sides of a pair therefore share the task, checkpoint
state, screenshot, and history, differing only in the recovery action. A pair's raw
margin is $\Delta_g = S_g(a^{+}_g) - S_g(a^{-}_g)$, and only groups with
$\Delta_g \geq 2$ are released, so every pair is separated by panel success count alone.
Groups below that threshold are dropped rather than resolved by a secondary signal:
success counts saturate when most of the panel recovers from every candidate, and we
prefer to discard those groups than to rest a preference on recovery efficiency. The
released set therefore contains no tie-broken pair, and its 100 targets are all
non-degenerate.

\paragraph{Composition of the released pairs.}
Table~\ref{tab:rm_composition} reports the outcome of selection. All five generators
appear frequently on both sides
(\subref{tab:rm_generator_distribution}); MAI-UI-8B is the most common negative and
Qwen3-VL-2B-Instruct the most common positive, but no single generator dominates either
role, and every generator appears at least 17 times in each. The mean margin is 2.86
panel successes and the maximum is $8/12$
(\subref{tab:rm_margin_distribution}), with half the pairs at the $\Delta_g{=}2$ release
threshold.

\begin{table}[h]
\caption{Composition of the 100 released preference pairs.}
\label{tab:rm_composition}
\centering\scriptsize
\begin{subtable}[t]{\linewidth}
\caption{Selected-action source distribution.}
\label{tab:rm_generator_distribution}
\centering\setlength{\tabcolsep}{9pt}
\begin{tabular}{lcccccc}
\toprule
& \multicolumn{5}{c}{Candidate generator} & \\
\cmidrule(lr){2-6}
Role
& \shortstack{Qwen3-VL\\2B-Instruct}
& \shortstack{Qwen3-VL\\8B-Instruct}
& \shortstack{GUI-Owl\\7B}
& \shortstack{GUI-Owl\\1.5-8B}
& \shortstack{MAI-UI\\8B}
& Total \\
\midrule
Positive & 22 & 19 & 19 & 20 & 20 & 100 \\
Negative & 19 & 22 & 17 & 19 & 23 & 100 \\
\midrule
Total & 41 & 41 & 36 & 39 & 43 & 200 \\
\bottomrule
\end{tabular}

\end{subtable}

\vspace{6pt}

\begin{subtable}[t]{\linewidth}
\caption{Panel-success margin distribution.}
\label{tab:rm_margin_distribution}
\centering\setlength{\tabcolsep}{6.5pt}
\begin{tabular}{lccccccc@{\hskip 12pt}c}
\toprule
$\Delta_g$ & 2 & 3 & 4 & 5 & 6 & 7 & 8 & Total \\
\midrule
Pairs & 50 & 30 & 13 & 3 & 1 & 1 & 2 & 100 \\
\bottomrule
\end{tabular}

\end{subtable}

\end{table}

\clearpage

\subsection{Reward Model Training}

\paragraph{UI Genie RM}
We train three reward models with the UI Genie preference data from the same Qwen3-VL-8B-Instruct checkpoint: two BT models that learn a scalar reward with a pairwise
preference loss, one on 64K and one on 128K pairs, and a generative SFT model that predicts a single
correct/incorrect classification token with next-token cross-entropy. Training
data for all three comes from the initial, UI-Genie-RM-517k release,
using the context variant that retains the last five interaction steps; each
example therefore carries at most five past screenshots, so that sequence length
and image budget stay bounded and identical across objectives. From this
corpus we draw 64K and 128K chosen--rejected pairs, spanning 82,960 and 95,498
unique referenced images, after removing every pair that overlaps the UI Genie
evaluation set; the 128K set contains the 64K set. BT consumes these pairs directly and is trained
to assign the higher scalar reward to the chosen action; SFT expands each of the 64K pairs
into its two sides, giving 128K pointwise rows labeled with a correct/incorrect
token. Gradient accumulation is doubled for SFT so that SFT and the 64K BT model observe
the same 64 pairs per optimizer step, holding the visual contexts and actions
fixed across objectives. The 128K BT model keeps every other setting and trains for
twice as many steps, so the two BT models differ only in data scale.
Table~\ref{tab:baseline_rm_hparams} reports the complete
training configuration.

\begin{table}[h]
\centering\scriptsize\setlength{\tabcolsep}{5pt}
\renewcommand{\arraystretch}{1.15}
\caption{Full training configuration for the BT (64K and 128K pairs) and generative SFT baselines.}
\label{tab:baseline_rm_hparams}
\begin{tabular}{@{}p{0.25\linewidth}p{0.21\linewidth}p{0.21\linewidth}p{0.21\linewidth}@{}}
\toprule
Hyperparameter & BT (64K) & BT (128K) & SFT \\
\midrule
\multicolumn{4}{@{}l}{\textit{Objective and readout}} \\
Base checkpoint & \multicolumn{3}{c}{Qwen3-VL-8B-Instruct} \\
Training objective
& \multicolumn{2}{c}{Bradley--Terry pairwise loss}
& Next-token cross-entropy \\
Loss scope
& \multicolumn{2}{c}{Chosen and rejected scalar rewards}
& Final judgment turn only \\
Training data & 64K pairs & 128K pairs & 128K rows (64K pairs) \\
\addlinespace
\multicolumn{4}{@{}l}{\textit{Adaptation}} \\
Training method & \multicolumn{3}{c}{LoRA (rank 64, $\alpha=32$, dropout 0.05)} \\
LoRA target modules & \multicolumn{3}{c}{All linear LM layers} \\
Additional trainable parameters
& \multicolumn{2}{c}{Scalar score head}
& LM output head \\
Frozen vision encoder & \multicolumn{3}{c}{Yes} \\
Frozen vision--language aligner & \multicolumn{3}{c}{Yes} \\
Trainable parameters & \multicolumn{3}{c}{174.59M} \\
\addlinespace
\multicolumn{4}{@{}l}{\textit{Optimization}} \\
Training epochs & \multicolumn{3}{c}{1} \\
Per-device batch size & \multicolumn{3}{c}{1} \\
Gradient accumulation steps & 8 & 8 & 16 \\
Effective global batch size & 64 pairs & 64 pairs & 128 rows (64 pairs) \\
Learning rate & \multicolumn{3}{c}{$1\times10^{-4}$, cosine schedule, warmup ratio 0.05} \\
Optimizer & \multicolumn{3}{c}{Fused AdamW, weight decay 0.1} \\
Total optimizer steps & 1,000 & 2,000 & 1,000 \\
\addlinespace
\multicolumn{4}{@{}l}{\textit{Systems}} \\
Maximum sequence length & \multicolumn{3}{c}{8,192} \\
Maximum image pixels & \multicolumn{3}{c}{1,048,576} \\
Numeric precision & \multicolumn{3}{c}{bfloat16} \\
Attention implementation & \multicolumn{3}{c}{FlashAttention} \\
Distributed training & \multicolumn{3}{c}{DeepSpeed ZeRO Stage 3} \\
Hardware & \multicolumn{3}{c}{8$\times$ NVIDIA A40 (single node)} \\
\bottomrule
\end{tabular}
\end{table}

\clearpage
\section{Offline RL}
\subsection{AndroidFlux Offline RL Dataset for error recovery}
\label{app:offline_rl_data}

The reinforcement learning corpus supplies the decision points a policy is optimized on,
drawn from the same replayed AndroidFlux trajectories as the benchmarks. Each prompt
restores a device state together with the history that led to it and the goal still to be
satisfied. Construction proceeds in three stages: checkpoints are restored at four positions
around the first annotated error, prompts whose goal disagrees with their source trajectory
are removed, and every remaining prompt is annotated with whether its state already
completes the task. Training sets are then sampled from the resulting corpus.

\paragraph{Dataset Generation.}
Source trajectories come from ten agents executing AndroidWorld tasks: Qwen3-VL-2B, -4B,
-8B and -32B, GUI-Owl-7B, GUI-Owl-1.5-2B, -8B and -32B, Mobile-Agent-v3-7B, and MAI-UI-8B. As in the reward-modeling benchmark, task instances are produced by the AndroidWorld task generation pipeline, which instantiates each task template with randomly sampled parameters. The corpus covers 90 AndroidWorld root tasks.
 
Trajectories in which the detector--verifier pipeline (Appendix~\ref{detection_framework})
locates a first error supply the $t$, $t{-}1$ and $t{-}n$ prompts, and trajectories that the
AndroidWorld verifier marks successful supply the \emph{successful} prompts. A prompt pairs
the task goal with the executed history up to its checkpoint and the screen produced by the
last action. Each history step is rendered as one natural-language line---the source agent's
own action or summary text where it provides one, otherwise a deterministic narration of the
canonical action (Appendix~\ref{app:generalized-history})---so prompts from different source
agents share a single format. The checkpoint is the last action in the history, so a $t{-}1$ prompt
asks for the erroneous action, a $t$ prompt asks for the action at $t{+}1$, and no history
extends past the first error. Finally, each prompt's goal is compared with the goal recorded
at the first step of its source trajectory.

\paragraph{Checkpoint positions.}
Prompts are taken at four positions relative to the first annotated error $t$: at $t$, with
the error already committed; at $t{-}1$, one step before it, while the mistake is still
avoidable; at $t{-}n$, from a correct prefix well before it; and from trajectories that
reached the goal (\emph{successful}), which supply healthy decisions to balance the
error-derived positions. All twenty unseen-test task types are excluded.
Table~\ref{tab:rl_composition}\subref{tab:rl_position_distribution} gives the
distribution; each position contributes close to a quarter of the corpus.

\paragraph{Termination annotation.}
Every prompt is annotated with \texttt{is\_next\_terminate}, which marks states that
already satisfy the goal, so that the correct next action is \texttt{terminate}. A positive
label is assigned only when completion is established from the restored state; a negative
label means only that it was not.
Table~\ref{tab:rl_composition}\subref{tab:rl_terminate_distribution} reports the
positives: 212 prompts, 2.49\% of the corpus, most frequent in \emph{successful} and at
$t{-}1$ and rarest at $t{-}n$, where the error still lies well ahead.

\begin{table}[h]
\caption{Composition of the 8,499-prompt RL corpus. Shares in (b) are normalized within
each checkpoint position.}
\label{tab:rl_composition}
\centering\scriptsize
\begin{subtable}[b]{0.49\linewidth}
\caption{Checkpoint position.}
\label{tab:rl_position_distribution}
\centering\setlength{\tabcolsep}{4pt}
\begin{tabular}{llrr}
\toprule
Position & Checkpoint & Prompts & Share \\
\midrule
$t{-}n$    & correct prefix before the error & 2,097 & 24.67\% \\
$t{-}1$    & one step before the error       & 2,080 & 24.47\% \\
$t$        & error already committed         & 2,080 & 24.47\% \\
successful & goal-reaching trajectory        & 2,242 & 26.38\% \\
\midrule
Total      &                                 & 8,499 & 100.00\% \\
\bottomrule
\end{tabular}

\end{subtable}%
\hfill
\begin{subtable}[b]{0.43\linewidth}
\caption{Termination annotation.}
\label{tab:rl_terminate_distribution}
\centering\setlength{\tabcolsep}{4pt}
\begin{tabular}{lrrr}
\toprule
Position & Prompts & Terminate & Share \\
\midrule
$t{-}n$    & 2,097 &  17 & 0.81\% \\
$t{-}1$    & 2,080 &  69 & 3.32\% \\
$t$        & 2,080 &  49 & 2.36\% \\
successful & 2,242 &  77 & 3.43\% \\
\midrule
Total      & 8,499 & 212 & 2.49\% \\
\bottomrule
\end{tabular}

\end{subtable}

\end{table}

\clearpage
\subsection{Training Details}
\label{app:offline_rl_details}

\paragraph{Policy.}
All offline-RL experiments initialize from Qwen3-VL-4B-Instruct (revision \texttt{ebb281ec}).
We finetune the language-model component using LoRA with rank $r{=}64$ and scaling factor $\alpha{=}32$, applied to all linear layers.
The vision tower and vision--language aligner remain frozen.
For each training context, the policy input consists of the system prompt containing the tool schema, the task goal, the latest 10 history steps rendered as natural-language text, and the current screenshot.
The policy outputs a short reasoning and action description followed by a single tool call, with spatial
coordinates represented on a normalized $0$--$1000$ grid.

\paragraph{GRPO optimization.}
We use the GRPO implementation in \texttt{ms-swift}.
For every prompt, the policy samples $K{=}8$ candidate completions.
Each optimization step contains 16 prompts and therefore 128 sampled completions.
All runs contain exactly 2,400 prompts and are trained for 150 steps, corresponding to one epoch over the prompt set.
We use a learning rate of $5\times10^{-5}$, 5\% warmup, and cosine decay to $0.1$ times the initial learning rate.
The KL coefficient is $\beta=10^{-3}$.
Rollouts use temperature $0.9$, top-$p$ $0.95$, and a maximum completion length of 512 tokens.

\paragraph{Reward model.}
Rewards come from the frozen UI-Genie Bradley--Terry reward model trained on 128K balanced UI-Genie
pairs (BT-128K in Table~\ref{tab:rm_main}).
Its input is the task goal, the full executed history as text, up to the five most recent screenshots,
and the full policy completion (reasoning, action description and tool call) as the candidate.
Candidate coordinates are rescaled from the policy's $0$--$1000$ grid to the pixel resolution declared
in the reward model's system prompt, which is the space the reward model was trained on.

\paragraph{Group Preference Reward Shaping.}
Let $r_i$ denote the scalar reward-model score assigned to candidate $i$ in a group of $K=8$ samples.
Following GPRS~\citep{zhu2025gprs}, we define the group win share
\begin{equation}
B_i =
\frac{1}{K}\sum_{j=1}^{K}\mathbb{1}[r_i \ge r_j],
\end{equation}
and use the centered value
\begin{equation}
A_i = B_i - \frac{1}{K}\sum_{m=1}^{K} B_m
\end{equation}
as the advantage.
We do not divide by the within-group standard deviation.
An unparseable completion receives $B_i=0$, but remains in the group and therefore still acts as an opponent when computing the win shares of parseable candidates.

\paragraph{Terminate guard.}
Reward-model ranking alone can otherwise assign positive reward to a syntactically valid \texttt{terminate} action even when the current task has not yet been completed.
We therefore apply a termination guard after reward scoring.
For a candidate whose parsed action type is \texttt{terminate}, we set $B_i=0$ unless the current AndroidFlux context is labelled
\texttt{is\_next\_terminate=True}, indicating that the goal has already been satisfied and termination is the appropriate next action.
UI-Genie contexts do not contain this annotation; for these rows, we fall back to whether the reference action is a terminating action.
The guard changes only the shaped policy reward and does not modify the underlying reward-model score.

\paragraph{Terminal-state annotation.}
Every released AndroidFlux RL context contains an
\texttt{is\_next\_terminate} annotation specifying whether the task goal is already satisfied at that state, such that the correct next action is termination.
The labels are produced with Gemini-3.8-Flash using two agreeing passes, a strict acceptance gate, and consistency checks across neighbouring trajectory states.
These annotations are used by the termination guard during policy optimization.

\paragraph{UI-Genie pool.}
For generic GUI training states, we deduplicate the 64K balanced UI-Genie reward-model pairs into 42,304 prompts and draw a 12,000-prompt pool stratified by history length, reference action type and app domain (seed 42).

\paragraph{Training-mixture construction.}
All offline RL experiments use the same budget of 2,400 training prompts, while varying only the source distribution. We construct six training sets from generic UI-Genie contexts and AndroidFlux contexts at the first-error checkpoint $t$ and one step before the error $t{-}1$. Each source pool is shuffled once (seed 42) and every set takes the first $k$ prompts of each pool, so the sets are nested; the $t$-only set cycles the 2,080 available $t$ prompts and repeats 320 of them. No reward-model or rollout-based filtering is applied. The resulting training sets are summarized below.

\begin{center}
\small
\setlength{\tabcolsep}{6pt}
\begin{tabular}{lrrr}
\toprule
Training mixture & UI-Genie & AndroidFlux $t$ & AndroidFlux $t{-}1$ \\
\midrule
UI 100\%                  & 2400 & 0    & 0   \\
$t$ 100\%                 & 0    & 2400 & 0   \\
UI75\% / $t$25\%                & 1800 & 600  & 0   \\
UI50\% / $t$25\% / $t{-}1$25\%      & 1200 & 600  & 600 \\
UI40\% / $t$35\% / $t{-}1$25\%      & 960  & 840  & 600 \\
UI50\% / $t$35\% / $t{-}1$15\%      & 1200 & 840  & 360 \\
\bottomrule
\end{tabular}
\end{center}

Here, $t{-}1$ denotes the checkpoint immediately before the first erroneous action, whereas $t$ denotes the first-error checkpoint and therefore conditions the policy on the erroneous action when predicting the recovery at $t{+}1$.

\paragraph{Evaluation.}
We evaluate checkpoint 150 on the 65 replay-based, human-verified AndroidWorld tasks, with one episode per task and condition and greedy decoding.
In the \emph{clean} condition, each task begins from its original initial state.
In \emph{error $t$}, we replay the recorded source trajectory to the state containing the effect of its first error and ask the learned policy to recover and complete the task.
Success is determined by the AndroidWorld programmatic task checker.

\paragraph{Variance and scope of the mixture study.}
Each training distribution is run with a single seed and evaluated on 65 tasks per condition.
As a reference for run-to-run variability, two UI-only runs from an earlier training recipe that differed only in random seed differed by 2/65 tasks (3.1 points) on clean evaluation and 4/65 tasks (6.1 points) on error-$t$ evaluation.
We therefore interpret the main experiment through the consistent improvement of AndroidFlux-containing mixtures over UI-only, rather than treating individual pairwise differences as independently significant.
The four mixtures form a small grid intended to test whether generic and failure-centered states are complementary; they should not be interpreted as an optimized mixture ratio.

\clearpage
\section{Qualitative: Succeeded and Failed Error Recovery Trajectory}
\label{app:qualitative}

\begin{figure}[h]
\centering
\includegraphics[width=\linewidth,height=0.85\textheight,keepaspectratio]{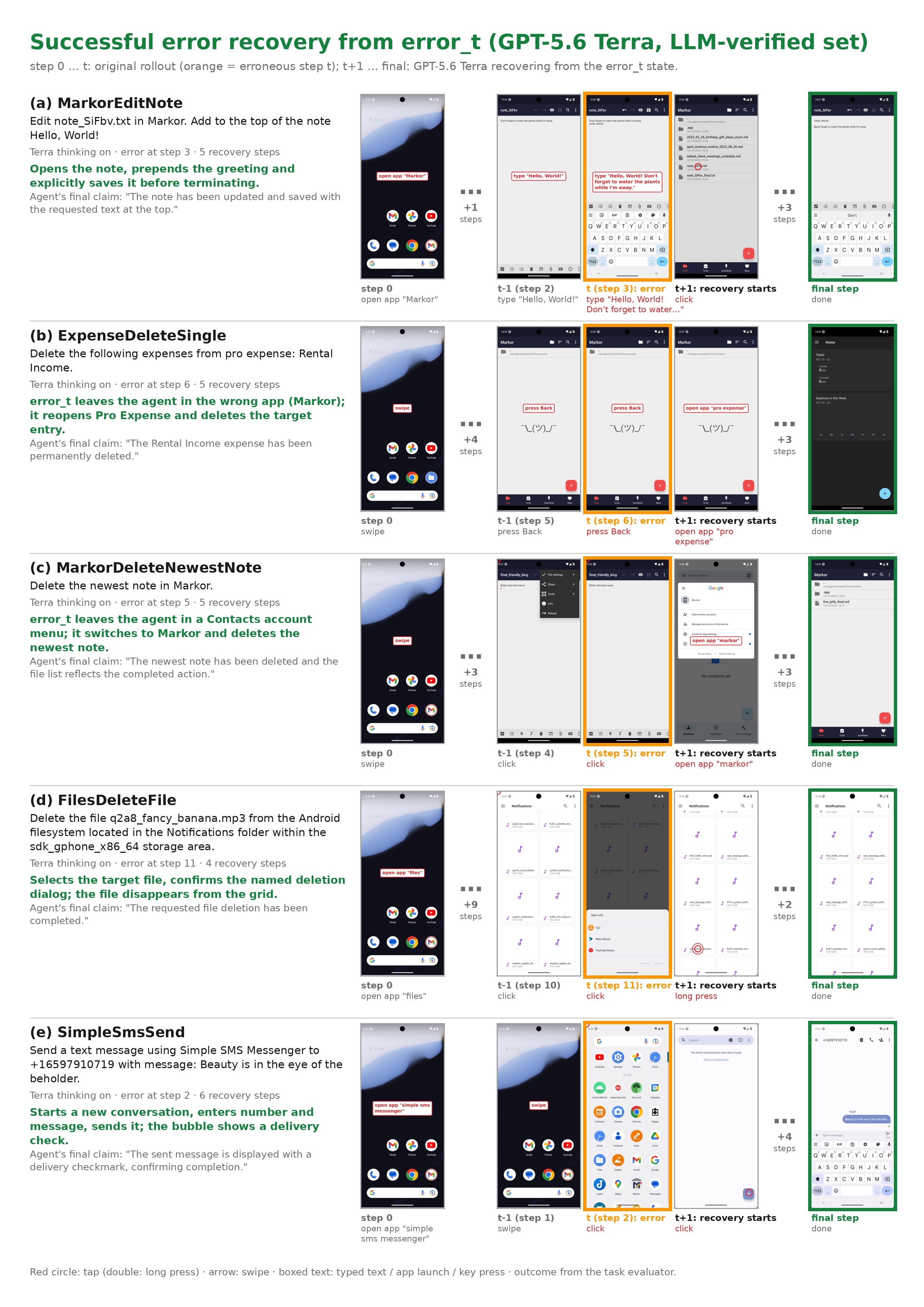}
\caption{Successful recovery trajectories from the first error at $t$ (GPT-5.6 Terra, think-on(medium), LLM-verified set).}
\label{fig:qual_success}
\end{figure}

\begin{figure}[h]
\centering
\includegraphics[width=\linewidth,height=0.85\textheight,keepaspectratio]{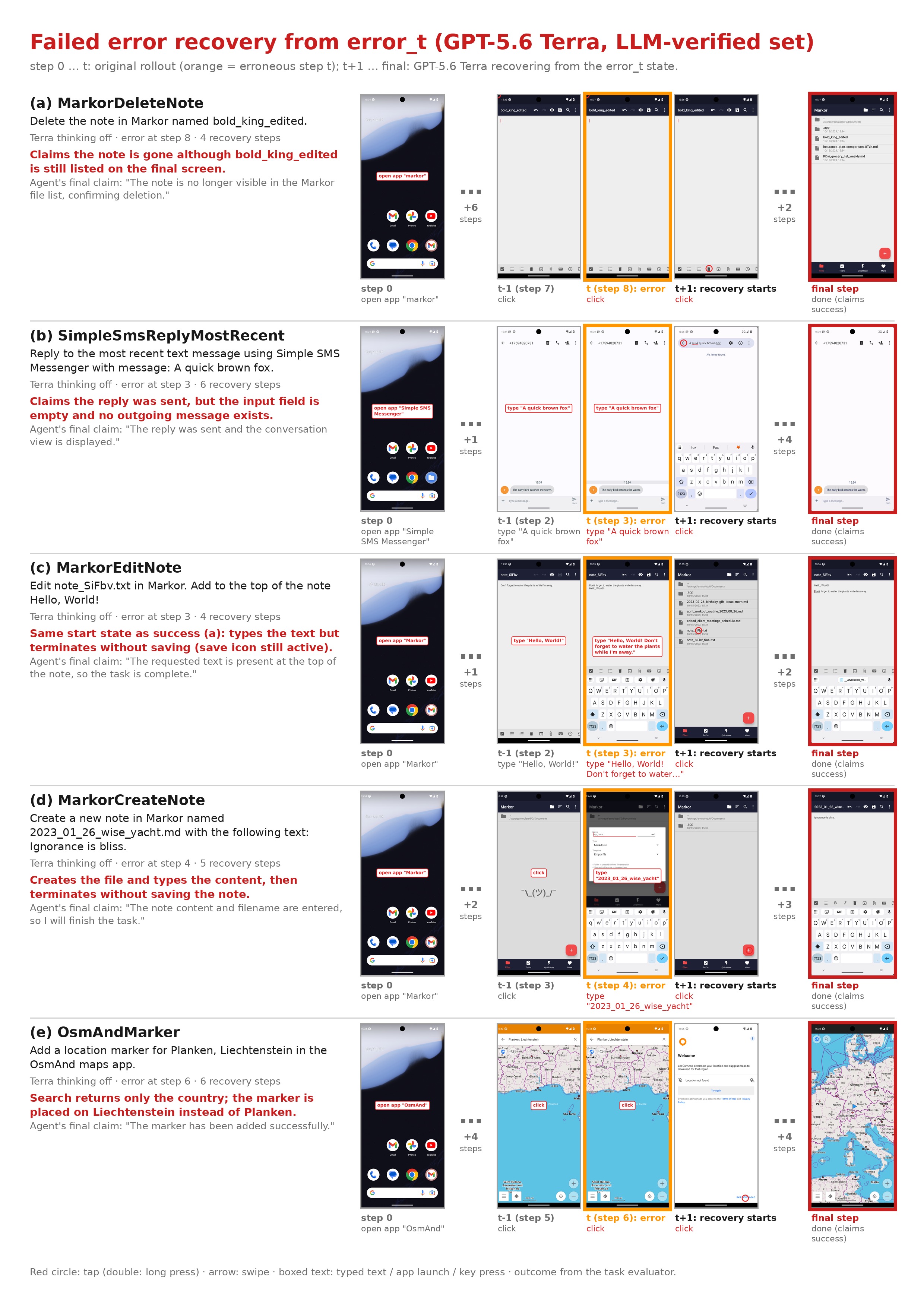}
\caption{Failed recovery trajectories from the first error at $t$ (GPT-5.6 Terra, think-on(medium), LLM-verified set).}
\label{fig:qual_failure}
\end{figure}

\clearpage
\bibliographystyle{plainnat}
\bibliography{androidflux}

\end{document}